\pdfoutput=1

\documentclass{article}
\usepackage{graphicx} 
\usepackage{amsmath}
\usepackage{microtype}
\usepackage{multirow,multicol}
\usepackage{pdflscape}
\usepackage{adjustbox}
\usepackage{listings}
\usepackage{mystyle}
\usepackage{caption}
\usepackage[utf8]{inputenc}

\usepackage{algorithm}
\usepackage{algpseudocode}
\usepackage{orcidlink}
\usepackage{authblk}   
\begin{document}
\bibliographystyle{unsrt}
\title{COOL: A Constraint Object-Oriented Logic Programming Language \\and its Neural-Symbolic Compilation System}
\author{\large Jipeng Han\orcidlink{0000-0002-9754-1000}\thanks{I am actively seeking a doctoral project opportunity. If my research interests align with your project and you would like to discuss potential collaboration, please feel free to contact me at \href{mailto:coolang2022@qq.com}{coolang2022@qq.com}. The ongoing project in this paper is hosted on GitHub and everyone is welcome to follow its evolution and contribute at \href{https://github.com/coolang2022/COOLang}{https://github.com/coolang2022/COOLang}.

} \\ \footnotesize Beijing Huagui Technology
}

\date{September 2023}

\maketitle

\section{Abstract}

\hspace{5mm} This paper explores the integration of neural networks with logic programming, addressing the longstanding challenges of combining the generalization and learning capabilities of neural networks with the precision of symbolic logic. Traditional attempts at this integration have been hampered by difficulties in initial data acquisition, the reliability of undertrained networks, and the complexity of reusing and augmenting trained models. To overcome these issues, we introduce the COOL (Constraint Object-Oriented Logic) programming language, an innovative approach that seamlessly combines logical reasoning with neural network technologies. COOL is engineered to autonomously handle data collection, mitigating the need for user-supplied initial data. It incorporates user prompts into the coding process to reduce the risks of undertraining and enhances the interaction among models throughout their lifecycle to promote the reuse and augmentation of networks. Furthermore, the foundational principles and algorithms in COOL's design and its compilation system could provide valuable insights for future developments in programming languages and neural network architectures.

\textbf{keywords:} COOL; Programming Language; Neural-Symbolic Compilation System; U2N; Neural-Symbolic Layer. 
\section{Introduction}
\hspace{5mm}
\begin{figure}[t!]
    \centering
    \makebox[\textwidth][c]{\includegraphics[width=1.2\linewidth]{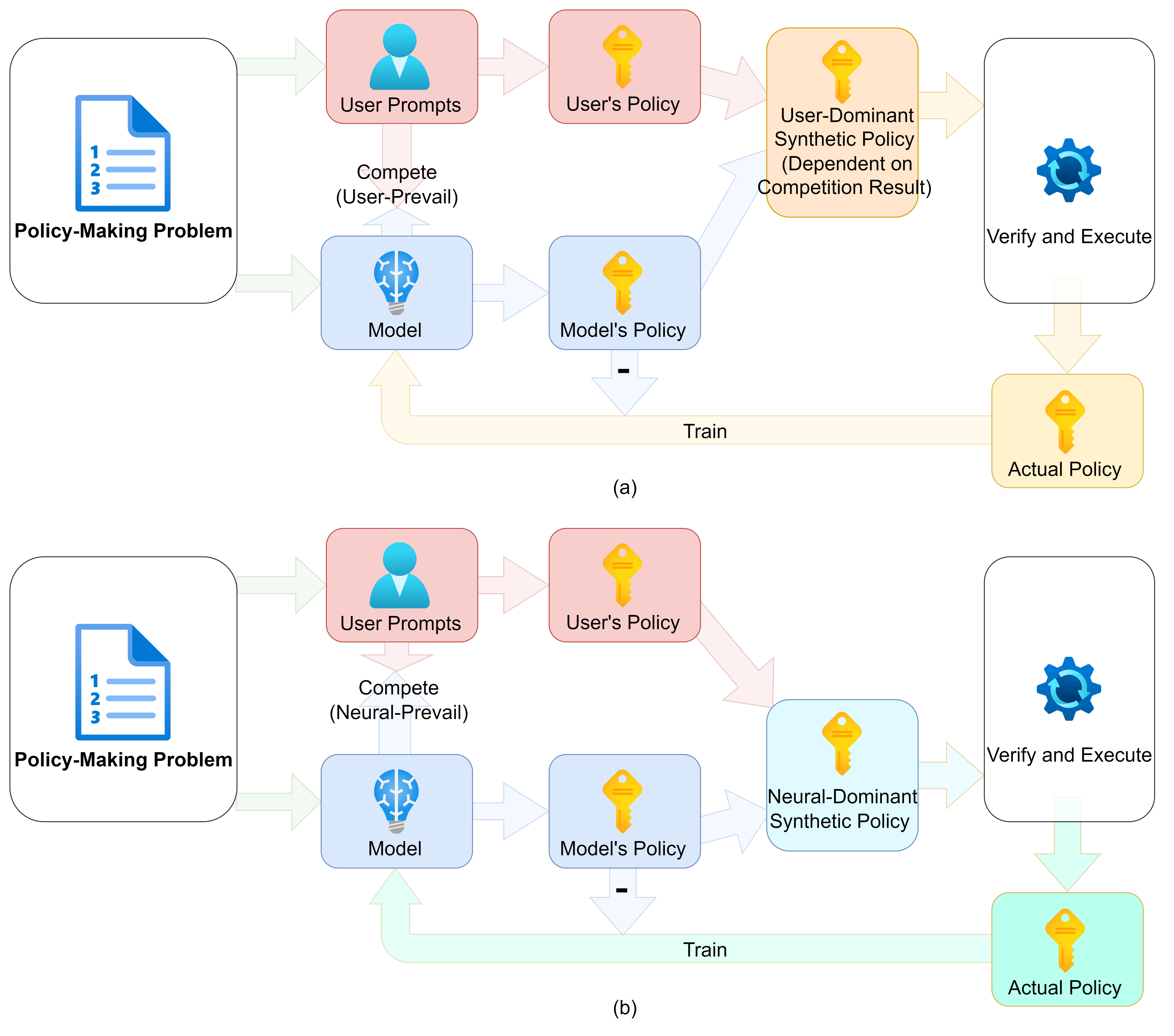}}
    \caption{The U2N mechanism}
    \label{fig:U2N_mechanism}
\end{figure}
Neural networks have displayed excellent learning and generalization capabilities across diverse areas, such as classification (e.g., image\cite{cai2020review_medical_image_classification} and speech recognition\cite{graves2014towards_speech_recognition}), sequential analysis (e.g., natural language processing\cite{belinkov2019analysis_nlp_survey} and time series forecasting\cite{lim2021time_time_series_forecasting_survey}), regression (e.g., predicting stock prices\cite{vijh2020stock_price_prediction}), generation (e.g., art\cite{santos2021artificial_visual_art_generation} and music creation\cite{ji2020comprehensive_music_generation}), etc. However, they often falter in calculation and formula manipulation. Therefore, the integration of neural networks and symbolic logical reasoning is important and promising, especially in cases that require both adaptability and precision, such as simulation, bioinformatics, fault diagnosis, and software engineering\cite{garcez2022neural_neural_symbolic_learning_survey}. 

Nevertheless, this integration faces significant challenges in the context of logic programming languages: 1) The scarcity of user and initial data for neural networks\cite{kumar2017programming_trend}, 2) The low effectiveness of integrating undertrained neural networks with logical reasoning, and 3) The difficulty in reusing and expanding these hard-won neural network models.

Previous efforts, including neural-guided logical reasoning\cite{ellis_learning_2018_neural_guided_bayesian,zhang_neural_2018_kanren_neural_guided_logc,dong_neural_nodate_neural_logic_mechines}, neural network logic regularization\cite{hu_harnessing_2016_regularization_1,wang_integrating_2020_regularization_2,winters_deepstochlog_2022_regularization_3,manhaeve_neural_2021_regularization_4,li_closed_2020_regularization_5,fischer_dl2_2019_regularization_6}, neural-logic program induction\cite{sen_neuro-symbolic_2022_induct_1,schmid_inductive_2018_induct_2,cropper_learning_2020_induct_5,shindo_differentiable_2021_induct_6,gao_learning_2022_induct_7}, and logical neural network templating\cite{weber_nlprolog_2018_neural_network_templating}, expand the aspects of neural network and logic programming integration but fail to pay enough attention to and address these practical issues: initial data acquisition, reliability in reasoning, and model reusability. This paper takes a different path from other studies by concentrating on solutions to these practical problems.

To address these challenges, this COOL is designed based on the User-to-Neural (U2N) mechanism: Neural networks learn from users, compete with users, and gradually take over the policy-making process. As illustrated in Figure \ref{fig:U2N_mechanism}, both the user and the neural network model contribute to the policy-making process. Initially, as depicted in part (a) of the figure, when a neural network model is undertrained, user prompts are predominant, and thus the resulting policy will be user-dominant. The compiler will reason following the guidance of this user-dominant policy. However, since the user's prompts may not be flawless and generalized, the policy that the compiler ultimately implements might not align completely with the initially formulated policy. By identifying and analyzing the differences between its proposed policy and the adopted one, the neural network models refine their strategies, better positioning themselves for subsequent rounds of competition. Leveraging their innate ability to learn and generalize, as the training progresses, neural network models begin to outcompete the user prompts in shaping the policies. This transition to a neural network-dominant policy is depicted in part (b) of the diagram. Throughout this evolutionary process, the shift in control from user-dominant to neural network-dominant in policy-making is gradual and localized.

The U2N mechanism offers several advantages over conventional approaches. First, in comparison to the Teacher-Student Model\cite{gou2021knowledge_teacher_student}, U2N eliminates the need for a specialized teacher, and its competitive mechanism enables the model to surpass the user, who acts as a teacher. Second, relative to Human-in-the-Loop\cite{wu2022survey_human_in_the_loop}, U2N requires the user to provide guidance only once, significantly minimizing interaction overhead. Finally, the U2N mechanism permits the immediate deployment of undertrained neural network models in a production setting, provided the entire system meets the requisite standards. Over time, the model will autonomously refine and enhance its capabilities.

\begin{figure}[t!]
    \centering
    \makebox[\textwidth][c]{\includegraphics[width=1.2\linewidth]{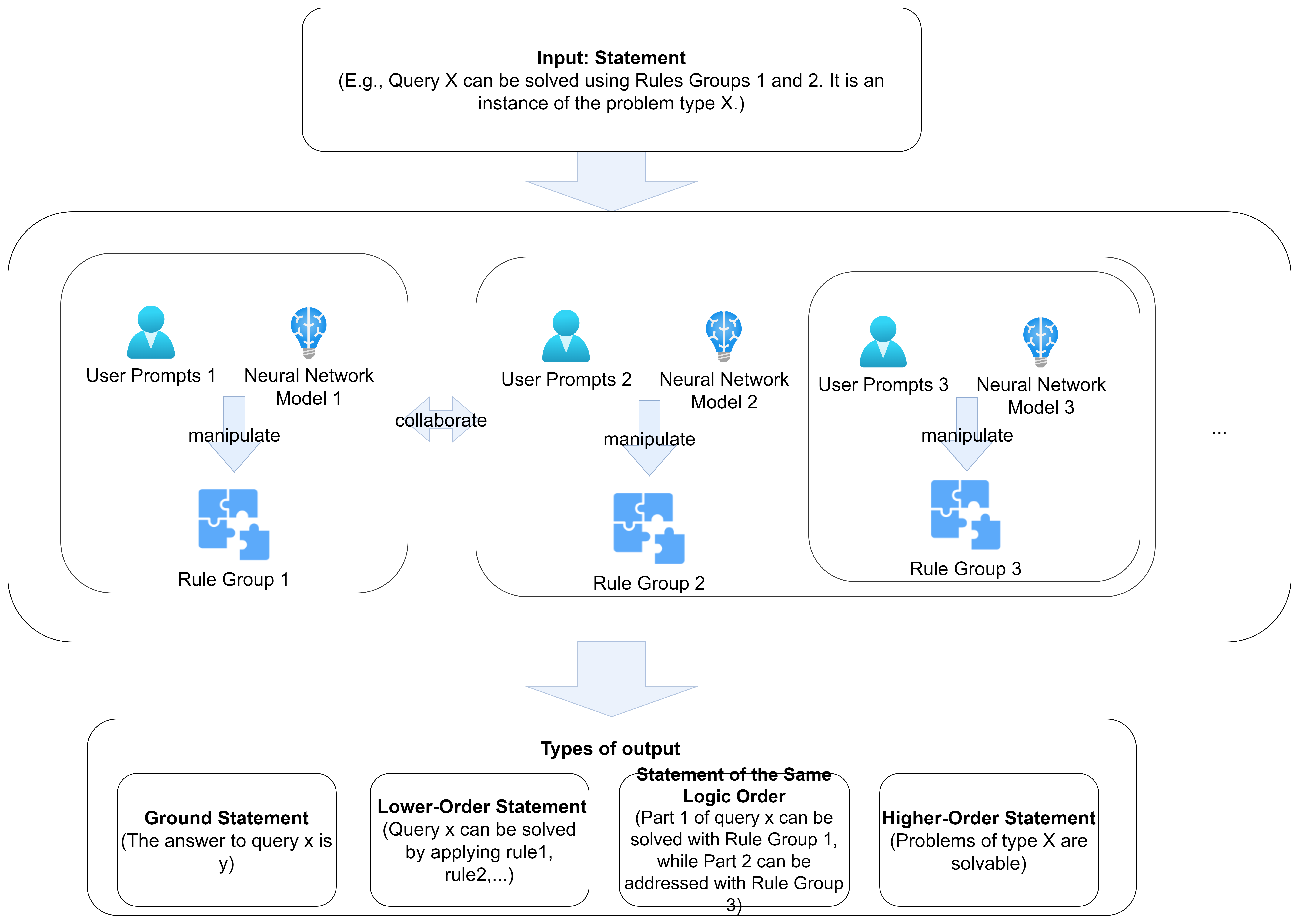}}
    \caption{Neural-Symbolic Layer}
    \label{fig:neural-symbolic-layer}
\end{figure}
\begin{figure}[t!]
    \centering
    \makebox[\textwidth][c]{\includegraphics[width=1.2\linewidth]{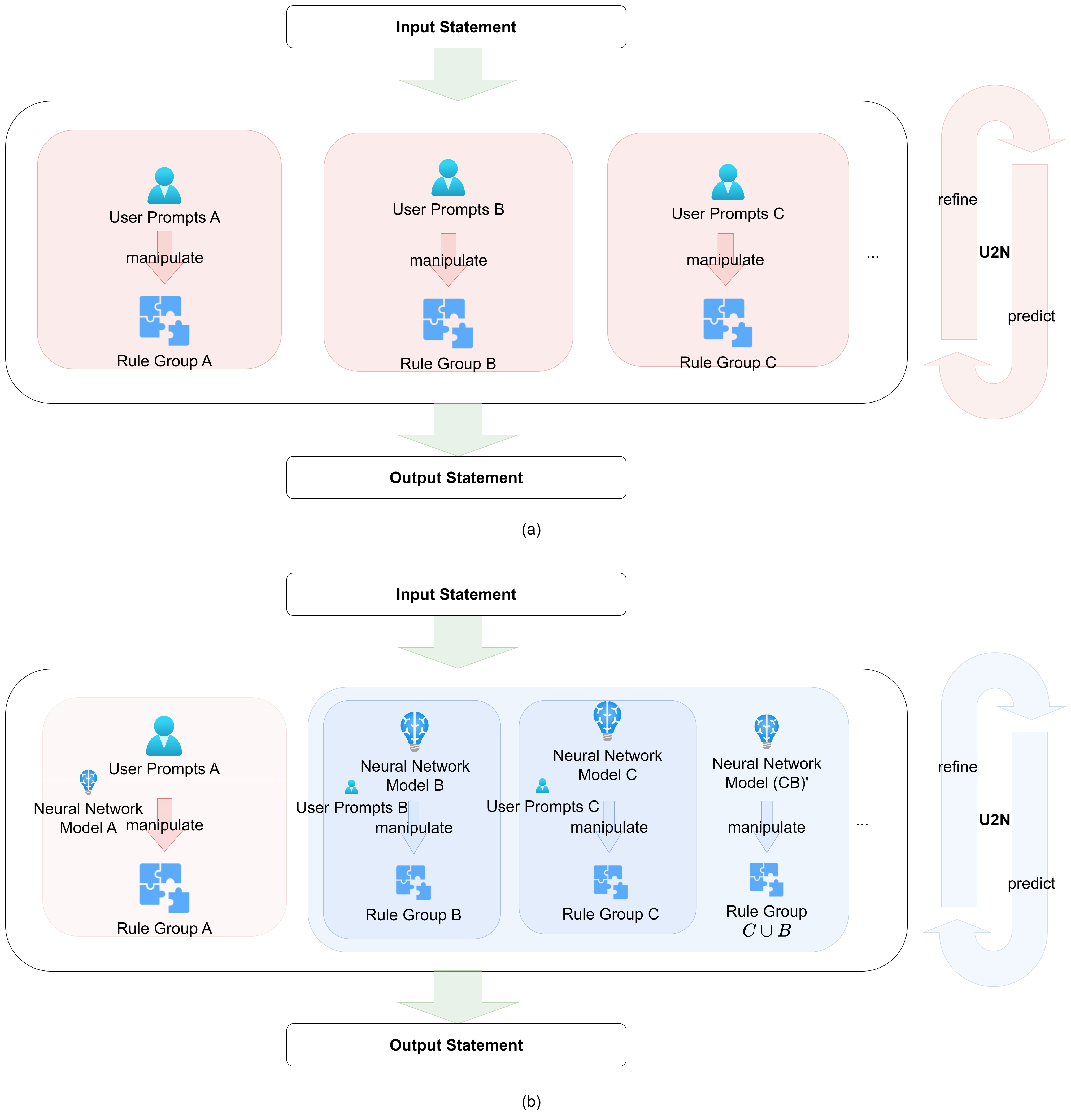}}
    \caption{Evolution of Neural-Symbolic Layer}
    \label{fig:nsl_u2n}
\end{figure}
Furthermore, as depicted in Figure \ref{fig:neural-symbolic-layer}, a novel layer for neural network architecture, termed the neural-symbolic layer, is proposed. This layer consists of rules grouped together and neural network models that manipulate them, aided by user prompts. Figure \ref{fig:nsl_u2n} illustrates the feedforward and backpropagation processes, which rely on both the U2N mechanism and the COOL compilation system's automated management of the entire lifecycle of neural network models, as well as their interactions with each other and the logic rules. Initially, as demonstrated in part (a), the layer possesses no neural network models. However, over time, as depicted in part (b), the layer adaptively develops neural network models that work on one or more rule groups based on their utilization. The training status of each neural network model is not synchronous. Neural-symbolic layers emphasize logical reasoning. Incorporating neural-symbolic layers into neural network models can substantially enhance the logical reasoning capability of the entire architecture.

To summarize, this paper contributes to:
\begin{itemize}
    \item \textbf{Programming Language Design:} COOL, a novel logic programming language with distinct features is designed.
    \item \textbf{Neural-Symbolic integration:} COOL's compilation system deeply integrates neural network and logical reasoning technology to achieve optimal coordination.
    \item \textbf{Code Synthesis:} COOL's compilation system uses a novel algorithm for efficient code synthesis during its reasoning process.
    \item \textbf{Guided Learning:} The U2N mechanism is a novel guided learning paradigm with its own advantages.
    \item \textbf{Neural Network Architecture:} Neural-symbolic layer is proposed and is promising to promote the logical reasoning ability of neural networks significantly.
\end{itemize}


\section{Related Work}

\subsection{Historical Overview}

\hspace{5mm}The integration of neural networks with logical reasoning has its roots in constraint logic programming. Since 1995, works like those by Bratko\cite{bratko_applications_1995_inductive_logic_programming} focused on the expressiveness of constraint logic programming to address inherent challenges. Over time, the focus shifted towards utilizing neural networks as pattern recognition modules in hybrid logic programming systems like DeepProblog\cite{manhaeve_neural_2021_regularization_4} and DeepStochLog\cite{winters2022deepstochlog_hybrid_stochastic}; or as prediction model to guide the reasoning process in logic programming system such as Neural Guided Constraint Logic Programming for 
 Program Synthesis (NGCLPPS)\cite{zhang_neural_2018_kanren_neural_guided_logc}, Neural Guided Deductive Search (NGDS)\cite{kalyan_neural-guided_nodate_neural_guided_search_synthesis_1}, Neural Program Search (NPS)\cite{polosukhin_neural_2018_neural_problem_solving_Neural_Program_Search3} and Neural-Symbolic Program Synthesis (NSPS)\cite{parisotto2016neuro_NSPS_4}. Our research is closer to the latter.

\subsection{Limitations of Existing Solutions}

\hspace{5mm}Solutions such as NGCLPPS, NGDS, NPS, and NSPS implement neural-symbolic integration largely adhering to the Powered By Example (PBE)\cite{lieberman2000programming_PBE,menon2013machine_PBE2} paradigm. A prominent limitation of the PBE paradigm is its dependency on the initial dataset, making it less suited for logical programming scenarios with scarce data. In addition, for a large part of them, as DSLs (Domain-Specific Languages), their scalability and reusability have also been challenged.

\subsection{Comparison with Existing Techniques}

\hspace{5mm}The COOL distinguishes itself from hybrid logic programming systems like DeepProblog, as well as neural-guided reasoning solutions like NGCLPPS, NGDS, NPS, and NSPS by leveraging the user's experiential knowledge in rule manipulation, automating the lifecycle of neural networks, and facilitating multi-model cooperation. Table \ref{tab:my_label1}, \ref{tab:my_label2}, and \ref{tab:my_label3} offer a comprehensive comparison of COOL against existing techniques:

\begin{table}[h]
    \centering
    
    \begin{tabular}{ccc}
    \hline
         Framework &  Base Language& User Contribution\\
         \hline
            DeepProbLog&  ProbLog\cite{de2007problog}&  Dataset; Reasoning Rules\\
          NGCLPPS&  miniKanren\cite{byrd2009relational_kanren}& Dataset\\
          NGDS&  PROSE DSL\cite{polozov_flashmeta_2015_PROSE}& Dataset\\
          NPS& LISP-inspired DSL & Dataset\\
          NSPS& FlashFill DSL\cite{gulwani2011automating_FlashFill,gulwani2016programming_Flash_fill} & Dataset\\
            COOL& COOL (language)& Reasoning Rules; Reasoning Prompts\\

         \hline
    \end{tabular}
    \caption{Overview of Base Language and User Contribution}
    \label{tab:my_label1}
\end{table}
 As illustrated in Table \ref{tab:my_label1}, The base language, COOL, is specially developed for the neural-symbolic system, aiming to provide a convenient environment for users to program with rules and prompts. This tailor-made language not only better realizes its purpose of making full use of user knowledge but allows for greater expansibility compared with DSLs or other adapted languages.

\begin{table}[h]
    \centering
    
    \begin{tabular}{cccc}
    \hline
         Framework &  Data Collection& Model Creation & Training, Testing and Updating\\
         \hline
            DeepProbLog&  Manual & Manual & Manual\\
          NGCLPPS&  Manual & Manual & Manual\\
          NGDS&  Manual & Manual & Manual\\
          NPS& Manual & Manual & Manual\\
          NSPS& Manual & Manual & Manual\\
            COOL& Auto& Auto &Auto\\

         \hline
    \end{tabular}
    \caption{Automation in Neural Network Model Pipeline}
    \label{tab:my_label2}
\end{table}
Table \ref{tab:my_label2} showcases the degree of automation implemented by different frameworks throughout the stages of the neural network model lifecycle, including data collection, model creation, and the cycles of training, testing, and updating. While frameworks such as DeepProbLog, NGCLPPS, NGDS, NPS, and NSPS rely on manual processes at each stage, COOL distinguishes itself by fully automating the entire pipeline. COOL’s comprehensive automation not only liberates users from repetitive tasks, enabling greater focus on development but also promotes consistency and improves the overall efficiency of managing neural network models.

\begin{table}[h]
    \centering
    
    \begin{tabular}{ccc}
    \hline
         Framework &  Multi-Model Prediction& Multi-Model Training\\
         \hline
            DeepProbLog&  Parallel & Parallel\\
          NGCLPPS&  Not Mentioned & Not Mentioned\\
          NGDS&  Not Mentioned & Not Mentioned\\
          NPS& Not Mentioned & Not Mentioned\\
          NSPS& Sequential  & Sequential \\
            U2N& Parallel; Composite & Contrastive \\

         \hline
    \end{tabular}
    \caption{Multi-Model Collaboration and Training}
    \label{tab:my_label3}
\end{table}
Table \ref{tab:my_label3} presents the strategies employed by various frameworks for multi-model prediction and training. An effective collaboration strategy is crucial for improving the reusability and adaptability of neural models across a range of tasks. The U2N framework adopts a dual approach: models operate in parallel for distinct parts of a task, while for intersecting segments, their predictions are combined into a composite outcome. In addition, U2N employs a contrastive training method, which enhances the ability of each model to identify its unique function within a task. This method enhances data utilization and the overall efficiency of collaboration.

\subsection{Advancements in Related Areas}
\hspace{5mm}
Neural-symbolic integration and hybrid systems have experienced significant advancements in recent years. In this section, we review key developments that are closely related to our work:

\begin{itemize}
    \item \textbf{Integration of LPLs and GPLs:} Several projects \cite{wielemaker_swi-prolog_2012_integration1, majchrzak_logic_2011_integration2, calejo_interprolog_2004_integration3} have explored the integration of Logic Programming Languages (LPLs) with General-Purpose Languages (GPLs). Inspired by these efforts, we have devised an approach to seamlessly combine these two language paradigms in the development of COOL, which features enhanced rule expressiveness and syntactic constructs that cater to GPL users.

    \item \textbf{Neural-Symbolic Reasoning:} Guiding logical reasoning with user prompts is a well-established practice. We have innovated by incorporating user prompts into the neural-symbolic reasoning process through the U2N mechanism, which is further integrated into the neural-symbolic layer designed for sophisticated logical reasoning.

    \item \textbf{Automation in Neural Network Model Lifecycle:} COOL is designed to promote a transition from user-driven to neural-driven processes throughout the neural network model lifecycle. This comprehensive approach encompasses automated data acquisition, model development, training, testing, evaluation, and deployment. In this context, COOL broadens the application scenarios of and the implementation methods for relevant technologies like Zero-Shot Learning \cite{pourpanah2022review_ZOL} and Self-Supervised Learning \cite{jaiswal2020survey_SSL}.

\end{itemize}

\section{Fundamentals}
  This section introduces the U2N mechanism and the Neural-Symbolic layer, two foundational concepts in the COOL's neural-symbolic compilation system.
\subsection{U2N Mechanism}
\hspace{5mm}
As depicted in Figure \ref{fig:U2N_mechanism}, under the U2N mechanism, the user and the neural network model have a mentoring and competitive relationship at the same time.
There are two primary attributes of the U2N mechanism: firstly, it permits the deployment of undertrained network models, and secondly, it enables neural network models to eventually surpass the user's performance, which acts as the initial teacher.

The key to achieving the first attribute lies in the competitive mechanism between the neural network and the user, ensuring that suboptimal policies are avoided and a minimum performance threshold is maintained post-deployment. This same mechanism also guarantees that the upper-performance limit is not restricted by user input. It is important to note, however, that while a neural network has the potential to eventually exceed user capabilities and tackle more intricate problems, its progression is incremental. As such, it is not designed to immediately address issues that are outside the scope of existing user policies. 

The foundation of the second attribute builds on the first: the model's ability to be deployed while undertrained and actively participate in the policy-making process allows it not only to learn from user strategy but also to adapt through direct environmental interaction, thus enabling the model to potentially surpass user expertise.

The User-to-Neural has two implications:
\begin{itemize}

    \item \textbf{User-to-Neural Knowledge Distillation}\\

    \hspace{5mm} Users store their knowledge in the form of prompts rather than data. These prompts implies users policy-making strategies. In each policy-making loop,  models actively learn from the strategies. This emphasis on strategy-centered learning, promises an artificial intelligence paradigm more aligned with human cognition.
    
    \item \textbf{User-to-Neural Dominance in Policy-Making} \\

    \hspace{5mm} The U2N policy-making process is initially user-dominated. this process becomes increasingly reliant on the neural network as it refines its capabilities. This change aligns human knowledge with machine potential effectively.

\end{itemize}

\subsection{Neural-Symbolic Layer}

\hspace{5mm}A \textit{neural-symbolic layer} is an architecture comprising multiple neural networks and an extensive set of grouped rules, where the neural networks manipulate the rules to facilitate formal transformations from one logical statement to another. As illustrated in Figure \ref{fig:neural-symbolic-layer}, a neural-symbolic layer can transform a logical statement into a ground statement (with no logical variable or undetermined variable, see in Section \ref{constructs_in_cool}), a lower-order statement, a statement of the same logic order, or a higher-order statement \cite{nadathur1998higher_logic_programming}. 

It is noteworthy that when addressing deductive problems, the layer should aim to produce a statement of a lower logical order; conversely, for inductive problems, it should strive to yield a statement of a higher logical order. If the logical order of a statement experiences fluctuations through multiple layers, the problem itself may undergo inequivalent transformations. While such transformations should typically be avoided when resolving inductive or deductive issues, they may be beneficial in generative tasks.

\subsection{COOL's Neural-Symbolic Compilation System}
\begin{figure}[t]
    \centering
    \makebox[\textwidth][c]{\includegraphics[width=1.2\linewidth]{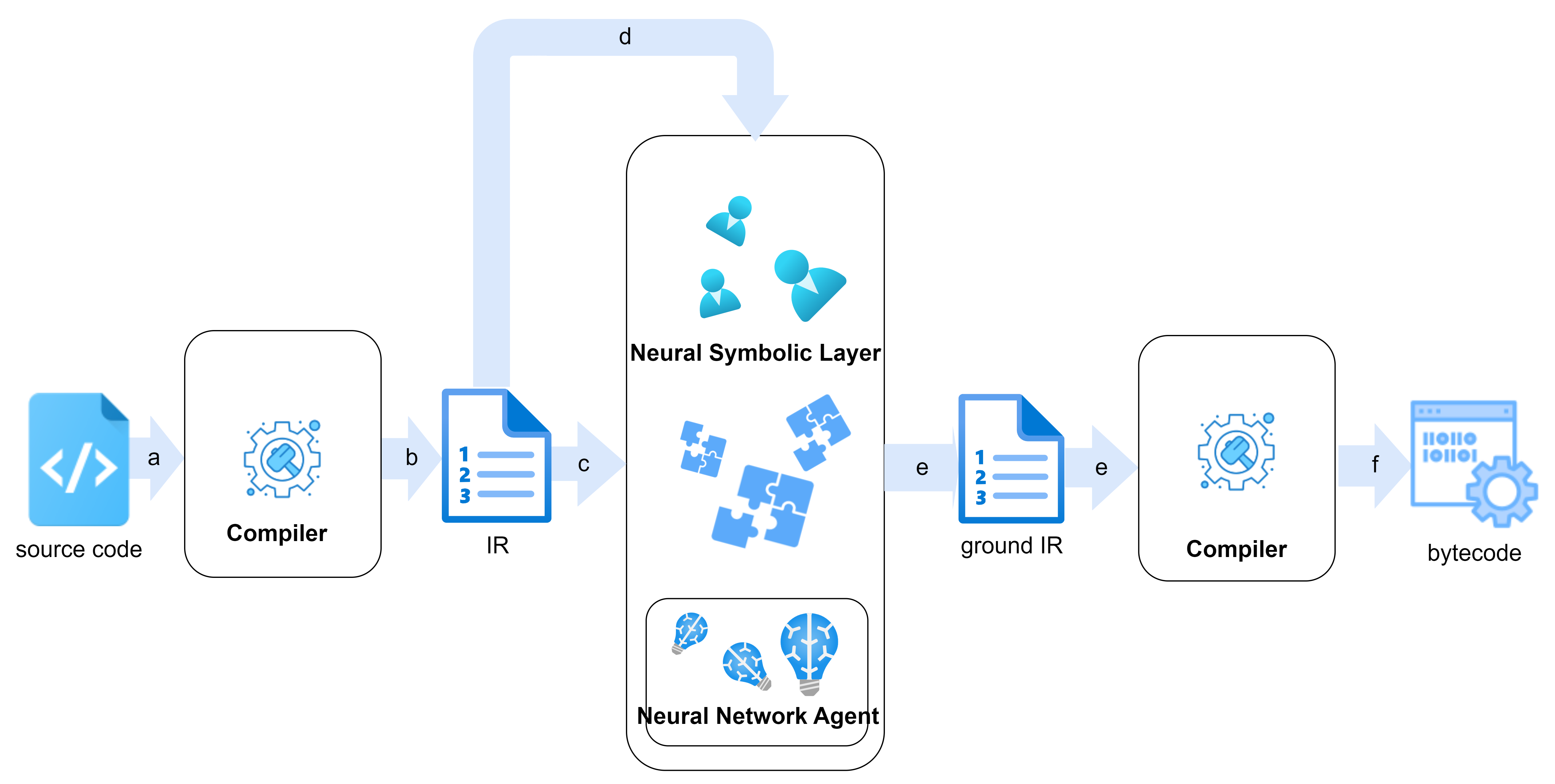}}
    \caption{Neural-Symbolic Compilation Process}
    \label{fig:Neural-Symbolic-Compilation-Process}
\end{figure}
\hspace{5mm} 
The COOL Neural-Symbolic Compilation System consists of a compiler and a neural network agent (introduced in Section \ref{sec:neural-network-agent}), which orchestrates the model management. Figure \ref{fig:Neural-Symbolic-Compilation-Process} illustrates the compilation flow from source code to bytecode, with steps a through f representing the sequence of data transformation:
\begin{enumerate}
    \item The compiler translates the source code into Intermediate Representation (IR), as shown in steps a and b.
    \item The IR, containing queries, rules, process control prompts, and domain-specific prompts (defined in Section \ref{sec:user_prompts}), is fed into the neural-symbolic layer. Each query is paired with domain-specific prompts that delineate the rule groups applicable for resolving the query, depicted in step c. Process control prompts, indicating user strategies for rule manipulation during reasoning, are sent to the neural-symbolic layer in conjunction with the rules to establish the neural-symbolic structure, as indicated in step d.
    \item The neural-symbolic layer translates the input IR into grounded IR (defined in Section \ref{sec:ground}), illustrated in steps c and e.
    \item Finally, the grounded IR is compiled into bytecode, as indicated by steps e and f.
\end{enumerate}

In this paper, while the COOL compiler is less central to the discussion and is not extensively covered, the logical reasoning components of COOL are thoroughly discussed in Section \ref{section:constraint_object_logic_programming}. The neural network agent, representing the neural network aspect, is detailed in Section \ref{sec:neural-network-agent}.

\section{COOL: Constraint Object-Oriented Logic Programming}
\label{section:constraint_object_logic_programming}
\hspace{5mm}
This section introduces COOL's language and compiler, including its structures, distinct paradigms, intermediate representation, approach to program synthesis, and the grounding process.
.
\subsection{Key Constructs in COOL}
\label{constructs_in_cool}
\hspace{5mm}The subsequent content delves into the core constructs of COOL. Each item will be discussed with a structured approach: beginning with a definition, followed by an elucidation rooted in code examples, and culminating in a summary to encapsulate the essence of the structure.

\begin{enumerate}
    \item \textbf{Fact Function:} 
    \begin{itemize}
        \item \textit{Definition:}
        A function that (1) produces either a simple value or has an implicit return, (2) includes at least one undetermined variable, and (3a) embeds undetermined variables within its name, termed an \textit{inverse function}, wherein the function body explicitly presents a determined value denoted as "ans" for its return, or (3b) designates the undetermined variable as its return, referred to as a \textit{forward function}.

        \item \textit{Example: (Code \ref{code:Fact_Function})} 
         Fact functions 1 and 2, although distinct in syntax, bear a resemblance to traditional GPL functions; notably, fact function 2 has an implicit return. Fact function 3 introduces an expression-based function name; the '\texttt{\$}' adorned 'x' signifies an undetermined variable in the equation $a+x == b$. This notation enhances the expressiveness of function names. Fact function 4, has a similar functionality to fact function 3, but resorts to the return "ans" to compute the value of x.
        \captionsetup[figure]{name=Code}
\begin{figure}[t!]
\begin{lstlisting}[language=C++]
@add(a)to(b){ //Fact function 1, forward function
    b=b+a;
}
@add(a,b){  //Fact function 2, forward function
    return:a+b;
}
@{a+$x==b}{  //Fact function 3, inverse function
    x=b-a;
}
@{a+$x}{  //Fact function 4, inverse function
    x=ans-a;
}
new:x=0;
1+$x==2;  //Call the fact function 3
\end{lstlisting}
\caption{ Fact Function }
\label{code:Fact_Function}
\end{figure}
\captionsetup[figure]{name=Figure}

        \item \textit{Essence:} Despite the structural resemblance to GPL functions, fact functions diverge significantly in mathematical characteristics and expression evaluation order. The concise mathematical definitions of the fact function and the algorithm for deducing the appropriate evaluation order are presented in Definition \ref{def:fact_function} and Algorithm \ref{alg:evaluation_order}, respectively. In the scope of this research, undetermined variables are treated as \textit{logic variables}, and expressions that include logic variables are categorized as \textit{queries}. Under this conceptual framework and within the realm of logic programming, when a fact function is invoked, it leads to the instantiation of logic variables within the body of the function. 
        Essentially, fact functions encapsulate the unification process in logic programming systems.

        \begin{definition}\label{def:fact_function}
        
                Let \( f \) be a fact function. Formally, \( f \) can be defined as:
                \begin{equation} f: \textbf{x} \mapsto y \end{equation}
                where \( \textbf{x} \) denotes a set of input variables derived from the function's name and \( y \) is the resultant value of the function. Specifically, the function adheres to one of the two distinct formulations:
                \begin{enumerate}
                    \item \( f: \textbf{x} \mapsto y \) where \( \textbf{x} \) consists only of determined variables. \( f \) is termed as the \textit{forward function}.
                    \item \( f^{-1}: (\textbf{x}_d, y) \mapsto \textbf{x}_u \) in which \( \textbf{x}_d \) and \( \textbf{x}_u \) represent the determined and undetermined parts of the input variables, respectively. \( f^{-1} \) is termed as the \textit{inverse function}.
                \end{enumerate}
        
        \end{definition}

        \begin{algorithm}[t!]
        \caption{Evaluation Order Deduction}
        \label{alg:evaluation_order}
        \begin{algorithmic}
        
        \Require An expression \( \mathcal{E} \) composed of fact function calls.
                \Ensure An ordered sequence of function calls for evaluation.
                
                \State \textbf{Parse} \( \mathcal{E} \) from left-to-right\cite{python_eval_order} to produce \( \mathbf{f} = [f_1, f_2, \dots, f_n] \).
                \State \textbf{Partition} \( \mathbf{f} \) into:
                \begin{align*}
                \mathbf{f}_{\text{forward}} & = [f_{\text{fwd,1}}, \dots, f_{\text{fwd,k}}] \\
                \mathbf{f}_{\text{inverse}} & = [f_{\text{ivs,1}}, \dots, f_{\text{ivs,l}}]
                \end{align*}
                with the constraint \( k + l = n \). Maintain relative order from \( \mathbf{f} \) within both subsequences.
                \State \textbf{Construct} the final evaluation sequence \( \textbf{f}^* \) as: 
                \begin{align*} \mathbf{f}^* = \textbf{f}_{\text{forward}} \parallel \text{reverse}(\mathbf{f}_{\text{inverse}}) \end{align*}
                In the context of this algorithm, \( \parallel \) denotes the concatenation operation and \(\text{reverse}\) is the operation to reverse the sequence.
                
                \Return \( \mathbf{f}^* \)

        \end{algorithmic}
        \end{algorithm}

    \end{itemize}

    \item \textbf{Rule Function:} 
    \begin{itemize}
        \item \textit{Definition:} A function returns an expression, distinctively marked by the 'expr' modifier. This function's return will be integrated at the point of invocation.
        
        \item \textit{Example (Code \ref{code:Rule_Function}):} In rule functions 1, 2, and 3, the variable $b$ is presented without a prefix, prefixed with \texttt{\$} or prefixed with \texttt{\#}, respectively. These denote $b$ as a determined variable, an undetermined variable, or a variable that could be either, in that order. While rule functions share similarities with macros or inline functions, they differ in invocation; they are not directly callable. Rule functions do support recursive structures, as exemplified in Rule function 4.

\captionsetup[figure]{name=Code}
\begin{figure}[t!]
\begin{lstlisting}[language=C++]
expr:@{ln(a*b)}{  //Rule function 1
    return:ln(a)+ln(b);
}
expr:@{ln(a*$b)}{  //Rule function 2
    return:ln(a)+ln(b);
}
expr:@{ln(a*#b)}{  //Rule function 3
    return:ln(a)+ln(b);
}
expr:@{(a) is the parent of (b) && (b) is the parent of (c)}{  //Rule function 4
    return: (a) is the parent of (b) && (b) is the parent of (c) && (a) is the grandparent of (c);
}
\end{lstlisting}
\caption{Rule Function}
\label{code:Rule_Function}
\end{figure}
\captionsetup[figure]{name=Figure}

        \item \textit{Essence:} In terms of logic programming, rule functions can be interpreted as functionally-represented rules.
    \end{itemize}
    
    \item \textbf{Constraint-Query Function Group:}
    \begin{itemize}
        \item \textit{Definition:} A construct comprised of two essential parts: (1) a comprehensive fact function with solely determined variables in its name, and (2) at least one declarative fact function name containing one or more undetermined variables.
        
        \item \textit{Example (Code \ref{code:Constraint_Query_Function_Group}):} The first part, referred to as the ``constraint component," clearly defines the constraints and relationships between various variables. This includes relationships such as the one between quantity, unit price, and total price, or between initial speed, launch angle, and the landing point's distance. On the other hand, the second part, named the ``query component", specifies which variables are to be queried. It's crucial that the names of these variables remain consistent with those in the constraint component. For instance, it determines the unknown quantity when the unit price and total price are fixed, or the undetermined initial speed when both the distance to the landing point and the launch angle are known. Significantly, these two distinct components can be invoked independently.

\captionsetup[figure]{name=Code}
\begin{figure}[t!]
\begin{lstlisting}[language=C++, escapeinside={(*@}{@*)}]
/*Constraint-query function group 1*/
@(a) kg of apples at (b) per kg costs{ //part 1, constraint component
    return:a*b;
}=>@($a) kg of apples at (b) per kg given costs; //part 2, query component
new:w = 0;
($w) kg of apples at (3) per kg given costs == 50; //call the part 2

/*Constraint-query function group2*/
@projectile distance with speed (v0) and angle ((*@\(\theta\)@*)){  //part 1, constraint component
     v0x=v0*cos((*@\(\theta\)@*));
     v0y=v0*sin((*@\(\theta\)@*));
     t=2*v0y/g;
     return:v0y*t;
}=>@speed ($v0) at angle ((*@\(\theta\)@*)) given distance;  //part 2, query component
new:v=0;
speed ($v) at angle ((*@\(\pi\)@*)/3) given distance == 1000;  //call the part 2

\end{lstlisting}
\caption{Constraint-Query Function Group  }
\label{code:Constraint_Query_Function_Group}
\end{figure}
\captionsetup[figure]{name=Figure}

        \item \textit{Essence:} The core capability of constraint-query function groups lies in their ability to facilitate inverse parameter computations\cite{sickel1979invertibility_pl0,abramov2002universal_inverse_pl1,yokoyama2008principles_inverse_pl2} based on return values. From a logical programming perspective, constraint-query function groups can be viewed as functionally-represented constraint sequences and query clauses.
    \end{itemize}
    
    \item \textbf{User Prompts:}
    \label{sec:user_prompts}
    \begin{itemize}
        \item \textit{Definition:} Syntactic constructs within COOL that serve as interfaces, enabling users to relay high-level computational instructions to the compiler. These are comprised of two categories: domain-specification prompts and process-control prompts.
            \begin{description}
            \item {Domain-Specification Prompts (DSP)}: Constructs demarcating the range of permissible functions for invocation, including the inheritance mechanisms in object-oriented programming and file inclusion protocols.
            \item{Process-Control Prompts (PCP)}: Constructs determining the circumstances under which a function can be invoked. These are represented by prefix arrays placed before function names (following the `@` symbol). \\
            \end{description}
        \hspace{5mm}To better elucidate the underlying mechanics of PCP within the COOL compile, we present the following formal mathematical definitions and principles:\\

                \begin{definition}\label{def:pcp}
                Let $\mathbf{pcp}$ be a vector, termed a Process-Control Prompt, given by:
                \begin{equation}
                \mathbf{pcp} = [pcp_1, pcp_2, \ldots, pcp_n]
                \end{equation}
                where $n$ signifies the total number of steps in the procedure.
                
                Each element $pcp_i$ in $\mathbf{pcp}$ is defined as:
                \begin{equation}
                pcp_i = 
                \begin{cases} 
                0 & \text{if the associated function is not invokable at step } i \\
                r_p & \text{if the associated function is invokable at step  } i \text{ with reward } r_p 
                \end{cases}
                \end{equation}
                where $r_p$ is a non-zero real number.
                \end{definition}
                    
                \begin{definition}\label{def:step}
                    Let $\boldsymbol{\tau} = \{\tau_1, \tau_2, \ldots, \tau_n\}$ be a set of steps, where each $s_i$ represents a distinct phase in the problem-solving process as dictated by a user. The ``step set'' of a function associated with $\mathbf{pcp}$, denoted as $\boldsymbol{\tau}_{p}$, is given by:
                    \begin{equation}
                    \boldsymbol{\tau}_{p} = \{\tau_i \mid pcp_i \neq 0\}
                    \end{equation}
                    Hence, if $pcp_i \neq 0$ for a given $\mathbf{pcp}$, the corresponding operation is permissible during the $i^{th}$ step.
                \end{definition}

                \begin{definition}\label{def:reward}
                    Let $R_p$ denote the reward vector corresponding to $\mathbf{pcp}$, with each $r_{pi}$ (or equivalently $pcp_i$ when non-zero) representing the reward associated with executing an operation at step $i$. Formally,
                    \begin{equation}
                    R_p = \{r_{pi} \mid r_{pi} = pcp_i \text{ and } pcp_i \neq 0\}
                    \end{equation}
                    A higher magnitude of $r_p$ (where $r_p$ is an element of $R$) indicates a stronger user preference for executing the associated operation.
                \end{definition}
                
                \begin{principle}\label{prin:pcp_operation}
                    Let $\tau_m$ be the current step in the problem-solving process, and $\boldsymbol{\tau}_p$ be the step set derived from a $\mathbf{pcp}$. A function bound with the $\mathbf{pcp}$ is executable if and only if:
                    \begin{equation}
                    \exists \tau_i \in \boldsymbol{\tau}_{p} : i \geq m
                    \end{equation}
                    
                    Define $r_{um}$ as the reward for executing this function at step $\tau_m$:
                    \begin{equation} 
                    r_{um} =  r_{ui},  \text{ where }  i = \min(\{k | \tau_k \in \boldsymbol{\tau}_p \text{ and } k \geq m\}) 
                    \end{equation}
                    After execution, update $\tau_m$ to:
                    \begin{equation}
                    \tau_m \leftarrow \tau_k, \text{ where } k = \min(\{k | \tau_k \in \boldsymbol{\tau}_p \text{ and } k \geq m\})
                    \end{equation}
                    
                    For a complete problem-solving journey, the terminal step must be included:
                    \begin{equation}
                    m_{final} \in \boldsymbol{\tau}_p, \text{ where } final = \max \{i | \tau_i \in \boldsymbol{\tau}_p \}
                    \end{equation}
                \end{principle}

\captionsetup[figure]{name=Code}
\begin{figure}[t!]
\begin{lstlisting}[language=C++, escapeinside={(*@}{@*)}]
#load(io);  //Domain-specification prompt
class: Quadratic Equation Solver{ 
    exp:@(-2,0,0){$a==b}{  //Process-control prompt
        return:a-b==0;
    }
    exp:@(0,1,0){#a-b}{  //Process-control prompt
                return:a+(-b);
    }
    @(0,0,2){ a*$x^2+b*x+c==0 }{  //Process-control prompt
        x=(-b+(b^2-4*a*c)^0.5)/(2*a);
    }
}
class : Main (*@<<@*) Quadratic Equation Solver{ //Domain-specification prompt
    new:x=1;
    1*$x^2+4*x==100;
};
Main:m;
m.x-->screen;

\end{lstlisting}
\caption{User Prompts}
\label{code:User_Prompt}
\end{figure}
        \item \textit{Example (Code \ref{code:User_Prompt}):} The provided code segment aptly demonstrates the implementation of DSPs and PCPs. DSPs, such as ``\#load(io)" and ``Main \texttt{<<} Quadratic Equation Solver" (indicating that ``Main" inherits the ``Quadratic Equation Solver"), delineate functions that are permissible for use. DSPs play an essential role in problem resolution and must be provided by the user during programming. 
        
        Consider the three PCPs in class ``Quadratic Equation Solver" :
        \begin{equation}
        \mathbf{pcp}_1 = [-2,0,0]
        \end{equation}
        \begin{equation}
        \mathbf{pcp}_2 = [0,1,0]
        \end{equation}
        \begin{equation}
        \mathbf{pcp}_3 = [0,0,2]
        \end{equation}
        
        Associated with the PCPs are three distinct operations:
        \begin{enumerate}
            \item $\mathbf{pcp}_1$ - Moving the right side of the equation to the left.
            \item $\mathbf{pcp}_2$ - Converting subtraction to addition.
            \item $\mathbf{pcp}_3$ - Resolving the equation in its standardized form.
        \end{enumerate}
        
        From $\mathbf{pcp}_1$, it is evident that the first preferred action is to transpose terms from the equation's left side to the right at the $1^{st}$ step with a reward of -2. The negative value suggest the user's intention to minimize or be cautious about this operation, because this operation only needs to be performed at most once. Performing it repetitively will make the equation more complex. 
        
        $\mathbf{pcp}_2$ signifies that during the second step, the user prefers converting subtraction operations into additions, as evidenced by a reward of 1. Although this step streamlines the process, it doesn't constitute the key to the solution.
        
        Conversely, $\mathbf{pcp}_3$ conveys that at the $3^{rd}$ step, there's a strong inclination to solve the equation in its standard form, as seen by the relatively high reward of 2.
        
        If the compiler has currently advanced to the $2^{nd}$ step, only the operation connected with $\mathbf{pcp}_2$ and $\mathbf{pcp}_3$ can be executed next, ensuring the user's intentions are respected in the solution process.

        \item \textit{Essence:} From the perspective of logic programming, user prompts are a high-level kind of constraint on rules. Domain specification prompts ascertain the rules feasible for the grounding process, and process control prompts elucidate the methodology these rules should follow during said process.
    \end{itemize}
\end{enumerate}


These foundational elements of COOL are pivotal in comprehending the advanced concepts discussed in the subsequent chapters.

\subsection{COOL's Programming Paradigm}
\hspace{5mm}
The COOL language combines elements of GPLs like C++ and Java while adopting dynamic typing similar to Python. It seamlessly merges imperative, declarative, procedural, object-oriented, and functional paradigms. This integration endows COOL with the modularity, flexibility, and ease of use commonly associated with GPLs. As shown in Code \ref{code:Fact_Function}, \ref{code:Rule_Function}, \ref{code:Constraint_Query_Function_Group}, and \ref{code:User_Prompt}
, COOL distinguishes itself from traditional logic programming languages, especially ones similar to ProbLog and its offshoots\cite{de2007problog,costa2008visual_visual_problog,jaffar1992clp_CLP}. 
Unlike these languages where rules and sequenced constraints prevail, COOL interprets relations and constraints as functions to center its paradigm.

The evolution of this paradigm is rooted in key requirements from users and reasoning.

\begin{itemize}
    \item \textit{User requirements}: A prevailing dilemma in the realm of logic programming, especially within constraint logical reasoning, is the dissonance between its theoretical design and its practical applications. While declarative rules and constraints are conceptualized to facilitate users' logical reasoning processes, the often opaque nature of rule interactions can lead novices to overlook rules or introduce flawed constraints. Thus, to effectively harness the capabilities of logic programming languages, users must deeply understand the nature of the problem and the steps involved in problem-solving. Such languages fall short in providing a gradient environment, where users can gradually familiarize themselves with, and master, the problem-solving process. Consequently, many gravitate towards more intuitive GPLs, reflected in the teaching process of ProLog\cite{sekovanic_challenges_2022}. After overcoming challenges with a GPL, the allure to revert to an LPL diminishes, especially as most LPLs are niche Domain Specific Languages (DSLs). This dynamic suggests a likely predilection among users to prioritize proficiency in one or several GPLs over specializing in diverse LPLs. Notably, COOL is designed to offer holistic support for users with varying preferences and proficiencies. Users can initially engage with it as a GPL and, over time, delve deeper to harness the principles of Logic Programming to address challenges.
\end{itemize}

\begin{itemize}
    \item \textit{Reasoning requirements}: Through the introduction of GPL-style functions, the expressiveness of constraints and rules is elevated. This allows users to formulate more intricate constraints and rules to represent and tackle problems. With this advanced descriptiveness, previously exclusive to GPLs, COOL becomes capable of reasoning through more complex issues. Besides, Opting for functions instead of clauses provides the flexibility to embed additional information into the rules, such as the PCPs, which act as function attributes. Incorporating PCPs streamlines the logical reasoning process. This enhancement not only improves scalability but also significantly reduces the potential of encountering infinite loops due to cyclical rules\cite{bol1991analysis_cyclic_rule}.



\end{itemize}

\subsection{Intermediate Representation}
\hspace{5mm}
\begin{table}
    \hspace{-15mm}
    \begin{tabular}{ccccccccc}
    \hline
     \multirow{2}{*}{Code Type}    & \multirow{2}{*}{LHS} & \multirow{2}{*}{RHS} &\multirow{2}{*}{Operator}  & \multirow{2}{*}{Result} & \multicolumn{4}{c}{Attribute Flags}\\
 & & & & & LHS& RHS& Operator&Result\\
     \hline
     1&&&&1&0&0&0&0 \\
    4&a&b&COMMA&2&2&2&2&0 \\
    4&add\_ARG\_to\_ARG\_ &2&CALL&ans&3&0&2&2 \\
    2&&&&1&0&0&0&0 \\
    6&1&100&@&1&0&1&2&0 \\
    1&&&&6&0&0&0&0 \\
    4&b&a&+&7&2&2&2&0 \\
    4&b&7&=&b&2&0&2&2 \\
    5&&&&&0&0&0&0 \\
    2&&&&6&0&0&0&0 \\
    6&1&6&&1&0&0&0&0 \\

     \hline    
    \end{tabular}
    \caption{Intermediate representation for the fact function 1 in Code \ref{code:Fact_Function} }
    \label{tab:ir}
\end{table}

In the domain of compiler design and intermediate representations (IR), COOL utilizes the three-address code (TAC) as depicted in Table \ref{tab:ir}. Historically prevalent in GPLs \cite{reiser1981compiling_tac}, TAC offers a structured and concise format adept at delineating complex operations. Its design facilitates ease in parsing and subsequent modifications. With the one-to-one correspondence of each TAC instruction to a node in the rules and constraint trees, TAC is an ideal intermediate representation for COOL and an optimal data format for further neural network modeling.

\subsection{Program Synthesis in COOL}
\hspace{5mm}
\label{sec:ground}
Program synthesis in the context of COOL involves the transformation of COOL's Intermediate Representation (IR) from a hybrid declarative-imperative paradigm to an exclusively imperative form. This methodology echoes the strategies employed in Structured Query Language (SQL) and Halide, wherein physical expressions are generated\cite{graefe1993volcano_sql} and specific schedules are determined\cite{ragan2012decoupling_halide}, respectively.

Program synthesis in COOL is tied to its \textit{grounding} process. Originating from the logic programming domain, in which 'grounding' denotes the conversion of terms with logical variables into concrete ground facts, the term takes on a nuanced meaning in COOL. Here, 'grounding' characterizes the transformation of a query into fact function invocations. A comprehensive description of this transformation process is provided in algorithm \ref{alg:ir}.

\begin{algorithm}[t]
\caption{Code synthesis}
\label{alg:ir}
\begin{algorithmic}

\State Let $IR$ be an array of instructions.
\State Let $IR_{segment}$ be a segment of the array $IR$ corresponding to a query.
\State A ground $IR_{segment}$ has each instruction bound with a fact function.
\While{ungrounded $IR_{segment}$ exists}
    \State $\mathbf{f} \gets$ \{invokable functions from DSPs\}
    \State $ground\_flag \gets \text{False}$
    \Repeat
        \State Apply $f_{rule}\in\mathbf{f}$ or bind $f_{fact} \in \mathbf{f}$ to $IR_{segment}$ based on \textbf{principle} \ref{prin:pcp_operation}
        \If{$IR_{segment}$ is ground}
            \State Update original $IR_{segment}$
            \State $ground\_flag \gets \text{True}$
            \State \textbf{break}
        \EndIf
    \Until{resources exhausted or $ground\_flag$ is True}
    \If{not $ground\_flag$}
        \State \textbf{Error Exit}
    \EndIf
\EndWhile
\State \textbf{Success Exit}

\end{algorithmic}
\end{algorithm}

Program synthesis is indispensable for COOL's execution efficiency. Unlike conventional logic programming languages, such as ProLog, which perform grounding during execution, COOL separates the grounding phase from the execution phase. Owing to COOL's design that gathers queries, DSPs, and PCPs throughout the programming phase, the reasoning process does not require additional query or instructional input from interaction with users during execution. Therefore, integrating the grounding process within the compilation phase becomes a logical step. This integration ensures that, during the actual execution of the program, the time complexity parallels the efficiency standards of GPLs. A similar approach is adopted in Answer Set Programming (ASP), which generates ground programs before solving, thus reducing the problem to a propositional level\cite{gebser2022answer_ASP}.

\subsection{U2N Guided Grounding Process in COOL}
\hspace{5mm}
The purpose of the grounding process is to translate queries into specific fact function invocations. This involves working with the Abstract Syntax Tree (AST). Initially, the AST is composed of nodes that are not bound to any fact function. The compiler's role is to iteratively apply rule functions to modify the AST structure and bind fact functions to specific subtrees. A function can be applied or bound to a specific subtree only when the name of the function shares the same structure as the subtree, and this operation affects all nodes in the subtree simultaneously. The ultimate goal is for all nodes to be bound with fact functions, with each node being bound to only one fact function. 

Within each iteration, the compiler faces a crucial decision: to which subtree within the AST should which function be applied or bound? In the context of IR, this decision pertains to associating specific functions with particular sub-segments of the IR.


Given that the grounding state of each IR segment is independent of its preceding state, the success of converting a generated IR segment state during grounding is unaffected by any previous state. This characteristic renders each iteration of the grounding process as a distinct Markov Decision Problem. Consequently, the entire grounding procedure can be described as a sequence of Markov Decision Processes (MDPs)\cite{puterman2014markov}. 

But in practice, four important issues make the use of MDPs less feasible:
\begin{itemize}
    \item \textbf{Future Reward Calculation:} Given the myriad of potential combinations of IR segments and functions, and the unfeasibility of remembering each one, the compiler can only determine possible actions for a subsequent state after observing the current IR segment's state. This makes it infeasible to calculate the cumulative reward for future actions.

    \item \textbf{Backtrack Mechanism Necessity:} During the grounding process, it's not uncommon for an IR segment to reach an undesirable endpoint. Employing a backtrack mechanism can prevent repeated mistakes and conserve resources that would otherwise be wasted on re-grounding.
    
    \item \textbf{User Constraints:} While states in grounding are inherently independent, constraints imposed by users on the application order, preferences, and aversions through PCPs indirectly induce sequences in specific grounding states. As a result, states with higher accumulated rewards from the initial phase, relative to those with lower rewards, tend to be nearer to the successful conclusion of the grounding process or are more likely to complete the grounding with reduced resource consumption. Utilizing this information can hasten the identification of feasible solutions. However, it might challenge the fundamental Markovian structure of the process.
    
    \item \textbf{Step Skipping in Grounding Process:} As indicated by principle \ref{prin:pcp_operation}, grounding in line with PCPs allows for step skipping. However, bypassing an essential step on occasion can make an IR state ungroundable. Simultaneously, jumping directly to a later step can sometimes produce a significantly larger reward compared to regular states. This can cause the compiler to allocate considerable resources, only to realize that the state is ultimately infeasible. Addressing this challenge necessitates an appropriate mechanism.

\end{itemize}

\subsubsection{Bi-Direction Discounted Backtracking}
\hspace{5mm}
To address these challenges effectively, the COOL compiler introduces the Bi-Direction Discounted Backtracking (BDDB) algorithm, an evolution from traditional MDPs. This algorithm takes cues from Eligibility Traces\cite{singh_reinforcement_1996_Eligibility_Traces} and Offline Reinforce Learning\cite{levine2020offline_rl} in terms of processing and modeling with historical data, but there are significant distinctions between them. In the algorithm's name, "Bi-Direction" emphasizes its consideration of both past and future states. "Discounted" points to the use of discount factors, and "Backtracking" indicates the algorithm's ability to revisit prior states. In the BDDB algorithm \ref{alg:bddb} presented, the notation \( q(s, a) \) is employed to denote a value associated with a state-action pair. This is not to be confused with the Q-value from Q-learning.

\begin{algorithm}[t]
\caption{BDDB Algorithm}
\label{alg:bddb}
\begin{algorithmic}[0]
\Require $S, A, T, r, \pi, \gamma, \lambda, l, q_{base}$

\State \textbf{Initialize:}
\State State Space \( S = \{s_0\} \) (initially, will expand as more states are discovered)
\State Action Space $A=\{a_1, \dots, a_n\}$
\State Transition Function $s_{t+1} = T(s_t, a_t)$
\State Stochastic Policy $\pi(a_t | s_t)$
\State Reward Function $r(s_t, a_t, s_{t+1})$
\State Action-Value Function 
\begin{equation}
    q^\pi(s_t, a_t) = 
    \begin{cases}
    \mathrm{E}_\pi\left[\sum_{k=0}^{l} \gamma^k r_{t+k+1} \mid s_t, a_t, k \leq l \right] + r(s_t, a_t, s_{t+1}) + \lambda q^\pi(s_{t-1}, a_{t-1}) & \text{if } t \geq 1 \\
    r(s_0, a_0, s_{1}) +\lambda q_{base}& \text{if } t = 0
    \end{cases}
\end{equation}
\State \textbf{Execute:}
\State Action-Value Space $Q:\{q^\pi_{t,a_t} \rightarrow(s_t,a_t)\}$ initialized with $(q^\pi(s_0, a_0), (s_0, a_0))$

\While{exit conditions not met} 
    \State Select $(s_t, a_t)$ with highest $q$ from $Q$
    \State Compute $s_{t+1}$ via $T$ and corresponding $q^\pi(s_{t+1}, a_{t+1})$ 
    \If{$s_{t+1}$ meets success exit conditions}
        \State Output optimal action sequence $[a_0, \dots, a_t]$ for starting state $s_0$
        \State Update dataset $\mathbf{D}$ for further modeling
        \State Terminate
    \EndIf
    \State Remove $(q^\pi(s_t, a_t),(s_t, a_t))$ from $Q$ and add all $(q^\pi(s_{t+1},a_{t+1}),(s_{t+1}, a_{t+1}))$
\EndWhile
\end{algorithmic}
\end{algorithm}

\subsubsection{Elucidation of the BDDB Algorithm}

The BDDB algorithm provides a structured approach for problem-solving in scenarios characterized by a vast or infinite state space. The core components of the algorithm are: 

\begin{itemize}
    \item \textbf{State Space ($S$)}: Starting with the initial state $s_0$, this space expands to incorporate states identified during the algorithm's run. For COOL's grounding, a state symbolizes an IR segment, with $s_0$ being the original IR segment without any instruction bound to a fact function.

    \item \textbf{Action Space ($A$)}: Comprising valid actions $a_i$, which symbolize potential decisions in any state. In the context of COOL's grounding, an action signifies the act of applying or binding a function to an IR sub-segment, hence the viable actions are constrained by the state $s$.

    \item \textbf{Transition Function ($T$)}: This manages system development by mapping a state-action pair $(s_t, a_t)$ to the succeeding state $s_{t+1}$. Within COOL's grounding framework, the Transition Function equates to the compiler.

    \item \textbf{Policy ($\pi$)}: This stochastically maps actions to states, steering the system's operations. In COOL's grounding, the system employs batch processing for policy creation. Upon receiving an IR segment and a set of accessible domains $\textbf{d}$, which comprises accessible file and class names defined by domain specification prompts (DSPs) to constrain the feasible action range, the neural network agent responds with policies and two coefficients:
    \begin{equation} (\Pi,ac,ci) = \text{agent}(s,\textbf{d}) \end{equation}
    \begin{enumerate}
        \item Policy array $\Pi$ is defined as $\Pi = [\pi_1,\pi_2, \ldots,\pi_n]$, where $n$ signifies the IR segment's length and $\pi_i$ denotes the likelihood that a function can be applied or bound to the sub IR segment rooted at the node specified by the $i^{th}$ IR segment instruction. Notably, function specifics at sub-segments are not determined at this stage, rendering $\pi_i$ a general policy for all suitable potential actions.
        \item $ci$ represents the confidence with which the agent perceives the IR segment to be within its knowledge domain, specifically in relation to $\textbf{d}$.
        \item $ac$ denotes the agent's accuracy of neural network predictions, particularly in relation to the neural network's performance on the test set.
    \end{enumerate}
Compared to related frameworks, this processing approach considerably enhances efficiency. In works such as \cite{zhang_neural_2018_kanren_neural_guided_logc, kalyan_neural-guided_nodate_neural_guided_search_synthesis_1}, the neural network model must inspect the synthesized codes to evaluate them. Subsequently, it employs the neural network agent to iteratively score all produced codes. This method incurs significant computational overhead, with a complexity of $O(k \times n \times m)$, where $k$ signifies the average actions executed for a grounding/code generation process, $n$ represents node count, and $m$ indicates average rules/operations applicable to a node. In contrast, COOL's neural network omits the need to inspect actual $a_t$ and $s_{t+1} = T(s_t,a_t)$, leveraging $s$ and $\mathbf{d}$ directly to produce $\pi$ for each node of $s$ in batches. Consequently, its complexity is a mere $O(k \times m)$.

    \item \textbf{Reward Function ($r$)}: 

This function provides immediate feedback subsequent to the transition from state $s_t$ to $s_{t+1}$ post the execution of action $a_t$. Within COOL's grounding, the U2N mechanism in the compiler operationalizes the reward function. Based on this mechanism, immediate rewards are derived from both the user and neural network agent, transitioning from user-dominant to neural-dominant. The reward function in grounding is given as:
\begin{align}
r(s, \pi , r_p) &= (1-ac * ci) * ((r_p-r_a) * (1-ac* ci) +r_a) \\
&\quad + ac* ci* \pi * r_a + o
\end{align}

    with the following components:
    \begin{enumerate}
        \item $r_{p}$ is the reward provided by the associated process control prompt (PCP).
        \item $r_a$ is the reward from the neural network agent, used to counteract the user's guidance effect on the grounding process and impose the neural network agent's guidance effect:
        \begin{equation} r_a = \max \{k_0* |r_p| , r_{a_{base}}\}, \text{where } k_0 \in (0,1], r_{a_{base}} \geq 0 \end{equation}
        \item $o$ is an offset value. It serves as a reward for advancement aligned with steps specified by the PCPs or a gradually increasing penalty for stagnation:
        \begin{equation} o=
        \begin{cases}
            k_0 &\text{if } t=0\\ 
            -k_1*k_2^t & \text{if } t>0
        \end{cases}
        \text{where } k_0, k_1, k_2 > 0
        \end{equation}
    \end{enumerate}

\item \textbf{Action-Value Function ($q$)}: Denotes the anticipated cumulative reward when selecting action $a_t$ in state $s_t$ and adhering to policy $\pi$ subsequently. This function amalgamates immediate and future rewards and embodies the ``memory" concept by accounting for the value of prior state-action pairs. Two constants, $\gamma \in [0,1]$ and $\lambda \in [0,1]$, serve as discount and decay factors, respectively. Specifically, $\gamma$ attenuates the influence of future rewards, while $\lambda$ progressively lessens the impact of past rewards. In the COOL’s grounding context, $\gamma = 0$. This indicates that the compiler does not factor in future rewards, primarily because it cannot preemptively ascertain such rewards by bypassing time steps. On the other hand, with $\lambda \in (0,1]$, the compiler permits previously accrued rewards to inform and guide the grounding process. This approach harnesses the latent sequential information inherent in the grounding state. Moreover, by allowing the influence of past rewards to wane progressively, the method mitigates potential risks associated with skipping steps.

    \item \textbf{Action-Value Space ($Q$)}: This stores the relation between value and state-action pairs. It aids in pinpointing actions leading to maximum anticipated rewards and offers a mechanism to backtrack when a promising state-action pair leads to a less favorable outcome.

\item \textbf{Execution}: The algorithm, during its execution, explores the state-action space iteratively to uncover the optimal action sequence from the initial state $s_0$. The steps include:

\begin{enumerate}
    \item Choosing the state-action pair in $Q$ with the highest $q$ value.
    \item Ascertain the subsequent state via the transition function and calculate its associated action-value.
    \item If state $s_{t+1}$ meets predetermined success criteria, the algorithm renders the optimal action sequence from $s_0$. For COOL's grounding, the output is state $s_{t+1}$.
    \item Update $Q$ by excluding the currently evaluated state-action pair and incorporating the newly computed pairs.
    \item Exploration creates a dataset, \textbf{D}, which documents pertinent data from explored state-action pairs. This data can serve future modeling.
\end{enumerate}

In the context of COOL's grounding process, the arrays 
\begin{equation}
[(s_0, \text{root}(a_0),(\delta_\pi \mid (s_0,a_0)) ),\dots, (s_n, \text{root}(a_0),(\delta_\pi \mid (s_n,a_n)))]
\end{equation}
and 
\begin{equation}
\textbf{d}^{\prime} \mid [a_0,...,a_n]
\end{equation}
are assembled as modeling data $g$. Here, the function $\text{root}$ extracts the instruction position in $s$, where the instruction serves as the root of the sub-segment upon which the action operates. The policy error, denoted by \begin{equation}
    \delta_\pi = 1- \pi
\end{equation}
acts as a correction signal during modeling. The effective domain set $\mathbf{\mathbf{d}^\prime}$, a subset of $\textbf{d}$, represents the union of domains encompassing functions employed in $[a_0,...,a_n]$.

\item \textbf{Specific Scenarios}:

\begin{itemize}
    \item For $\lambda = 0$, BDDB mimics MDPs.
    \item For $\lambda = 1$ and $\gamma = 0$, BDDB acts as Breadth-First Search (BFS) with consistent negative rewards and as Depth-First Search (DFS) with positive rewards.
    \item When $ci*ca = 0$, user dominance prevails in the grounding process.
    \item At $ci*ca = 1$, the grounding process only adheres to the steps prescribed by the PCPs based on principle \ref{prin:pcp_operation}, with all other guidance being provided by the neural network agent. In this case, it is recommended not to disable PCPs to retain the user's most basic control over the grounding process.

\end{itemize}
\end{itemize}

    \section{Neural Network Agent}
    \label{sec:neural-network-agent}
\hspace{5mm}This chapter details the neural network agent, including data collection by the COOL compiler, model creation, training, testing, updating, and the prediction process.

\subsection{Data Collection}
The data collection is required by the neural network agent but integrated into the COOL compiler. When data collection is enabled by default via the settings file, each grounding process becomes a potential data source. 
\subsubsection{Procedure}
\begin{itemize}[leftmargin=0mm]

 \item \textbf{Data Generation:}
During each grounding process in the compilation, the compiler generates modeling data denoted as \(g\). 

 \item \textbf{Domain Grouping:}
Modeling data with the same effective domain set are grouped into the same knowledge domain. Both the effective domain set and corresponding knowledge domain are denoted as \(\textbf{d}^{\prime}\).
 \item \textbf{File Storage:}
Data from the same knowledge domain, generated within a given collection cycle (e.g., a week), are stored in a single file. The file is defined as \(\textbf{F} = \{ g \mid \textbf{d}^{\prime} \in g \text{ and all } g \text{ have the same value for } \textbf{d}^{\prime}\}\).

 \item \textbf{Indexing:}
Each file \(\textbf{F}\) is indexed by its associated \(\textbf{d}^{\prime}\).

 \item \textbf{Table Updating:}
The relationship between these indices \(\textbf{d}^{\prime}\) and their corresponding files is documented in table \(I\). The table is defined by \(I: \{\textbf{d}^{\prime} \mapsto \mathcal{F}\}\), where \(\mathcal{F} = \{\textbf{F}_t \mid t \text{ indicates the serial number of a collection cycle, } t \in \mathbb{N}\}\).

 \item \textbf{Notifying the Agent:}
After compilation, updates to the table \(I\) are relayed to the neural network agent for processing.

\end{itemize}

\subsubsection{Considerations}
\begin{itemize}[leftmargin=0mm]
\item \textbf{Privacy:} Users desiring confidentiality can modify settings to inhibit data gathering, offering them substantial discretion over their compilation datasets.
\item \textbf{Data Integrity:} Sourced directly from the compilation process, the compiler affirms the genuineness of the datasets. This direct sourcing obviates the necessity for supplementary verification post-collection.
\item \textbf{Granular Recording:} Datasets are meticulously categorized based on specific knowledge domains, contrasting with a broader classification by project. This differentiation ensures that varying problem-solving strategies are not conflated.
\item \textbf{Data Storage:} Datasets from distinct collection cycles are stored separately. Users possess the option to erase previously gathered data to conserve storage space.
\item \textbf{Incremental Modeling:} During multiple modeling sessions within a single collection cycle, only the most recent segment of the dataset is deemed as $new$.  Previously utilized data within current or earlier cycle is considered $old$ data, promoting a more streamlined and updated learning process.

\end{itemize}

\subsection{Construction of Composite Datasets}
\hspace{5mm}
Upon receiving the information about the generation of new data from compilations, the neural network agent undertakes the task of creating datasets for training and testing. Constructing datasets holds paramount significance for the neural network's performance. The dataset amalgamates new data with historical data. Concurrently, datasets corresponding to different knowledge fields are negatively sampled from each other for contrastive training \cite{jaiswal2020survey_contrastive}. 

In this section, unless explicitly stated, all sets are standard sets devoid of duplicated elements. When a multiset merges with another multiset or set, duplicate elements persist.

\subsubsection{Construction of the Positive Sampling Training Dataset}
\hspace{5mm} A modeling dataset, $\textbf{F}$, is partitioned into a training subset ($\textbf{F}^{train}$) and a testing subset ($\textbf{F}^{test}$) based on a predefined ratio during its initial utilization for modeling. The positive sampling training dataset, denoted as $\textbf{G}^{train}_{\textbf{d}^{\prime}}$ (a multiset), is formulated as:
\begin{equation}
\textbf{G}_{\mathbf{d}^\prime}^{train} = (\textbf{G}_{\mathbf{d}^\prime}^{train,new} \cup \textbf{G}_{\mathbf{d}^\prime}^{train,old}) \mid \{\textbf{F}_{\mathbf{d}^\prime}^{train}\}
\end{equation}

The constituents of this dataset include:
\begin{enumerate}
    \item \textbf{New Training Dataset} ($\textbf{G}_{\mathbf{d}^\prime}^{train,new}$): This multiset includes the most recent data produced by the compiler. Its inclusion in the training process is crucial. On one hand, this fresh data can help improve the performance of models that previously lacked sufficient training. On the other hand, this dataset reflects the most recent concept drift\cite{lu2018learning_concept_drift}, encompassing updates in knowledge domains—such as the modification of corresponding files and classes—and addressing new queries stemming from emerging real-world requirements that the neural network must adjust to. The formulation of $\textbf{G}_{\mathbf{d}^\prime}^{train,new}$ incorporates a re-sampling method (specifically, oversampling) \cite{more2016survey_resample} to enhance data variability and representation. Notably, this re-sampling approach doesn't prioritize data by weight; instead, it's grounded in their policy error, $\delta_\pi$.
    
    For $(s, \operatorname{root}(a),(\delta_\pi \mid (s, a))) \in \textbf{F}_{\mathbf{d}^\prime}^{new,train}$, the frequency of duplication is determined as:
    \begin{equation}
    n_{oversample} = \max\left\{\lceil n_{max} \times \frac{\delta_\pi - \delta_{tolerance}}{1 - \delta_{tolerance}} \rceil, 0 \right\}
    \end{equation}
    where:
    \begin{itemize}
        \item \( n_{max} \) represents the maximum permissible oversampling count.
        \item \( \delta_{tolerance} \in [0,1) \) is the paramount policy error that does not instigate duplication.
    \end{itemize}
    Further, the new training dataset is constructed as:
    \begin{align}
    \textbf{G}_{\mathbf{d}^\prime}^{train,new} = &\bigcup_{i=1}^{\mid \textbf{F}_{\mathbf{d}^\prime}^{new,train}\mid} \{ n_{oversample,i} \times (s_i, \operatorname{root}(a_i), InDom_i) \\
    &\mid (s_i, \operatorname{root}(a_i), \pi_i) \in \textbf{F}_{\mathbf{d}^\prime}^{new,train}\}\nonumber
    \end{align}
    where $InDom_i$ signifies whether $(s_i, \operatorname{root}(a_i))$ is within the knowledge domain $\textbf{d}^{\prime}$. Being a positive sampling process, $InDom_i$ is invariably true.

    \item \textbf{Old Training Dataset} ($\textbf{G}_{\mathbf{d}^\prime}^{train,old}$): This dataset aggregates data samples from previous datasets in this knowledge domain. Incorporating this dataset pursues two essential objectives. Firstly, it aims to stabilize accuracy fluctuations. Relying exclusively on the new data could lead to significant variations in accuracy metrics, potentially compromising the agent's consistent performance. Drawing samples from historical data can mitigate such fluctuations. Secondly, it serves a crucial role in counteracting data drift\cite{lu2018learning_concept_drift}. While models focused on a narrow knowledge domain might be relatively immune to data drift, those spanning a broader or composite knowledge domain (a domain derived from multiple others) might be susceptible to biases when perpetually trained on unbalanced datasets. This ensures that the neural network sustains a balanced allegiance across the entirety of the knowledge domain. Additionally, the old training dataset integrates a temporal weighting re-sampling strategy, specifically undersampling, to recalibrate the representation of prior training data based on the elapsed time since their initial sampling and the present moment.
    
    The old training dataset, $\textbf{G}_{\mathbf{d}^\prime}^{train,old}$, is defined as:
    \begin{equation}
    \textbf{G}_{\mathbf{d}^\prime}^{train,old} = \bigcup_{t=m-l}^{m-1} \mathscr{G}_{\mathbf{d}^\prime,t}^{train,old}
    \end{equation}
    where:
    \begin{itemize}
        \item $m$ denotes the ordinal of the current collection cycle.
        \item $l$ represents the length of the sliding window, which is a positive integer indicating the span of permitted historical data.
        \item $\mathscr{G}_{\mathbf{d}^\prime,t}^{train,old}$ is the dataset undersampled from $\textbf{F}_t^{train,old}$.
    \end{itemize}
    The cardinality of $\mathscr{G}_{\mathbf{d}^\prime,t}^{train,old}$ follows:
    \begin{equation}
    \mid \mathscr{G}_{\mathbf{d}^\prime,t}^{train,old} \mid = \psi^{m-t} \min\{\mid \textbf{F}_m^{train,new} \mid, \mid \textbf{F}_t^{train,old} \mid\}
    \end{equation}
    where $\psi \in [0,1]$ stands for the Sequential Attrition Undersample Rate (SAUR). It quantifies the proportion of data retained from older datasets when merging them sequentially with the newest dataset.

\end{enumerate}

\subsubsection{Construction of the Negative Sampling Training Dataset}
For two training datasets, $\textbf{G}^{train}_{\textbf{d}^{\prime}_1}$ and $\textbf{G}^{train}_{\textbf{d}^{\prime}_2}$, they are mutually negative when $\textbf{d}^{\prime}_1 \cap \textbf{d}^{\prime}_2 = \varnothing$. 

The negative sampling training dataset, $\textbf{G}^{train}_{\bot,\textbf{d}^{\prime}}$, adheres to:
\begin{align}
\begin{cases}
 \textbf{G}^{train}_{\bot,\textbf{d}^{\prime}} &= \{ (s_i,InDom_i) \mid \exists (s_i, \operatorname{root}(a_i), \text{True}) \in \operatorname{dedup}\left(\bigcup \textbf{G}^{train}_{\textbf{d}^{\prime}_\bot}\right) \\
 &\text{  and } InDom_i = \text{False for negative sampling} \} \\
\mid \textbf{G}^{train}_{\bot,\textbf{d}^{\prime}} \mid &= \phi\mid \textbf{G}^{train}_{\textbf{d}^{\prime}}\mid
\end{cases}
\end{align}
where:
\begin{itemize}
\item $dedup$ is the deduplication procedure. 
\item $\textbf{d}^{\prime}_{\bot}$ delineates the mutually exclusive counterpart of $\textbf{d}^{\prime}$, signifying two distinct knowledge domains. 
\item $\phi$ represents the negative sample rate, determining the negative sample proportion relative to positive samples.
\end{itemize}
\hspace{5mm}Constructing a training dataset using negative sampling is essential for models that engage in multi-domain knowledge collaboration. This approach enhances the model's proficiency in differentiating tasks specific to its knowledge domain, thereby enabling it to concentrate more effectively on its own tasks during collaboration.
\subsubsection{Finalization of the Training Dataset}
The complete training dataset, denoted as \( \mathcal{G}^{train}_{\textbf{d}^{\prime}} \), is a union of both the positive and negative sampling training datasets:
\begin{equation} \mathcal{G}^{train}_{\textbf{d}^{\prime}} = \textbf{G}^{train}_{\textbf{d}^{\prime}} \cup \textbf{G}^{train}_{\bot,\textbf{d}^{\prime}} \end{equation}

\subsubsection{Construction of the Testing Dataset}
The process for constructing the testing dataset, \( \mathcal{G}^{test}_{\textbf{d}^{\prime}} \), mirrors that of the training dataset with the following modifications:
\begin{itemize}
    \item 1 Instead of \( \textbf{F}^{train} \), \( \textbf{F}^{test} \) is utilized, as indicated by the replacement of all superscripts labeled as \( train \) with \( test \).
    \item 2 During the construction of \( \textbf{G}_{\mathbf{d}^\prime}^{test,new} \), oversampling is not employed. Specifically, \( \textbf{G}_{\mathbf{d}^\prime}^{test,new} \) is determined by the following:
\end{itemize}
\begin{align}
\textbf{G}_{\mathbf{d}^\prime}^{test,new} = \bigcup_{i=1}^{\mid \textbf{F}_{\mathbf{d}^\prime}^{new,test}\mid} \{&(s_i, \operatorname{root}(a_i), InDom_i) \\
&\mid (s_i, \operatorname{root}(a_i), \pi_i) \in \textbf{F}_{\mathbf{d}^\prime}^{new,test} \text{ and } InDom_i = \text{True}\} \nonumber
\end{align}

\subsection{Dynamic Training, Testing and Updating}
\label{sec:train_test_update}
\begin{figure}[!ht]
    \centering
    \makebox[\textwidth][c]{\includegraphics[width=1.2\linewidth]{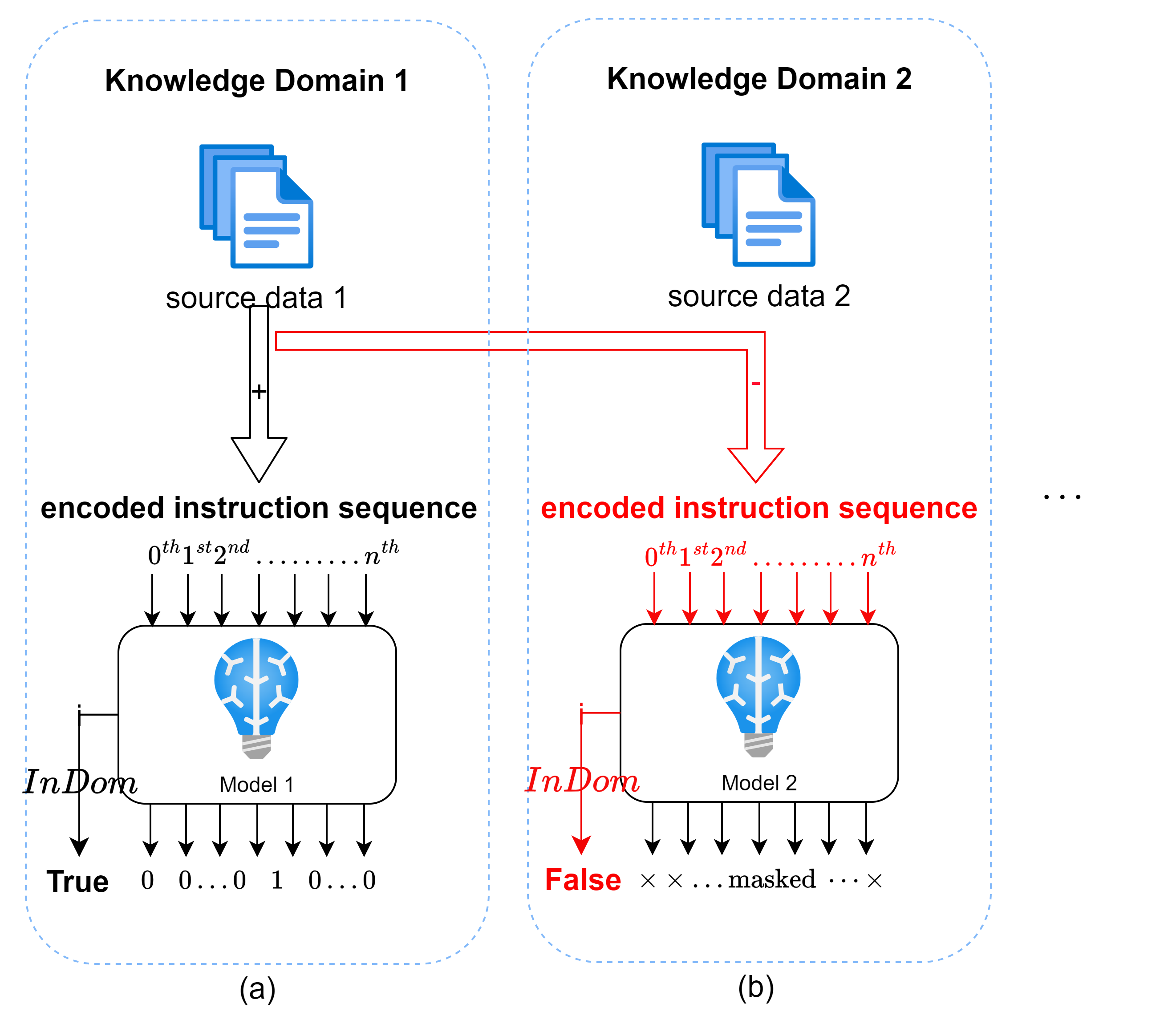}}
    \caption{Training}
    \label{fig:training}
\end{figure}
The constant evolution of programming tasks necessitates the continuous refinement of the U2N neural network models. With the composite datasets, the neural network agent undertakes the task of training, testing, and updating the models in the background, ensuring that they are always in sync with the most recent programming challenges.
    \subsubsection{Training}
As shown in figure \ref{fig:training}, the training process in U2N involves multi-model training and multitasking, and it includes a contrastive training process contingent on the compositions of the dataset. Consider a trained neural network model $M_{\mathbf{d}^{\prime},\theta}:{ e(s) \rightarrow (InDom, \Pi)}$, where $\mathbf{d}^{\prime}$ represents the knowledge domain to which the model is associated, $e(s)$ signifies the encoded form of instruction sequences, and $s$ denote a state in the grounding process. The function $e(s) \rightarrow (InDom, \Pi)$ describes the dual objectives of the model:  it takes a state $s$  in the grounding process as input, and outputs $InDom$, the confidence level that the state is within the grounding process pertinent to knowledge domain $\mathbf{d}^{\prime}$, and $\Pi$,  a policy array. Each element in $\Pi$ serves as an individual policy, indicating the probability of executing actions on sub Intermediate Representation (IR) segments that are based at a specific node.

Let the loss function for the first task be 
\begin{equation}
L_{InDom}(InDom_{\text{output}}, InDom_{\text{label}}) = \operatorname{BCE}(InDom_{\text{output}}, InDom_{\text{label}})
\end{equation}
where $\operatorname{BCE}$ denotes the Binary Cross-Entropy algorithm, and the loss function for the second task be 
\begin{equation}
L_{\Pi} = \operatorname{CCE}(\Pi_{\text{output}}, \Pi_{\text{label}})
\end{equation}
where $\operatorname{CCE}$ denotes the Categorical Cross-Entropy algorithm, and $\Pi_{\text{label}} = [\pi_i \mid i \in \mathbb{N}, i < |\Pi_{\text{output}}|, \pi_{i \neq \text{root}(a)} = 0 \text{ and } \pi_{i = \text{root}(a)} = 1]$.

In the training dataset $\mathcal{G}_{\mathbf{d}^{\prime}}^{\text{train}}$, a positive sample like $(s, \operatorname{root}(a), InDom = \text{True})$ can train both tasks, while a negative sample like $(s, InDom = \text{False})$ can only be used to train the first task. Therefore, the combined loss function is:
\begin{equation}
L = 
\begin{cases}
\epsilon * L_{InDom} + (1-\epsilon) * L_{\Pi}, & \text{if } InDom_{\text{label}} = \text{True}\\
\epsilon * L_{InDom}, & \text{if } InDom_{\text{label}} = \text{False}
\end{cases}
\end{equation}
where $\epsilon \in (0,1)$.

The training process is defined as:
\begin{equation}
\theta^* = \arg\min_\theta \bar{L} \,|\, M_{\theta}, \mathcal{G}_{\mathbf{d}^{\prime}}^{\text{train}}
\end{equation}

Here, $\bar{L}$ is the average loss and $\theta^*$ denotes the optimized neural network parameters.




    \subsubsection{Testing and Updating}

The performance of the updated model $M_{\theta^*, \mathbf{d}^{\prime}}$ is measured by its accuracies on two separate tasks:
The accuracy of predicting on \( InDom \) is defined as 
\begin{equation} 
A_{InDom} = \frac{\text{number of outputs where } | InDom_{\text{label}} - InDom_{\text{output}} | < 0.5}{\text{total number of outputs}} 
\end{equation}
and the accuracy of predicting on \( \Pi \) is defined as 
\begin{equation} 
A_{\Pi} = \frac{\text{number of outputs where } \pi_{\text{root}(a)} = \max(\Pi)}{\text{total number of outputs}} 
\end{equation}
Then, save the updated model along with its latest performance data \( A_{InDom} \) and \( A_{\Pi} \). Both the updated model and the accuracy reflect the improvements from the recent training iteration and will be used in the further guidance process.

It is important to note that if the dataset \( \mathcal{G}_{\mathbf{d}^{\prime}} \) contains no negative samples, then \( A_{InDom} \) will not be calculated nor updated. If \( M_{\mathbf{d}^{\prime}} \) has never been trained with a dataset containing negative samples, its \( InDom \) and \( A_{InDom} \) are considered invalid. When these two metrics are utilized, both are assigned the Jaccard similarity coefficient: 
\begin{equation} 
\frac{|\mathbf{d}^{\prime} \cap \mathbf{d}|}{|\mathbf{d}^{\prime} \cup \mathbf{d}|} 
\end{equation}

\subsection{Expand New Models}
\begin{figure}[!t]
    \centering
    \makebox[\textwidth][c]{\includegraphics[width=1\linewidth]{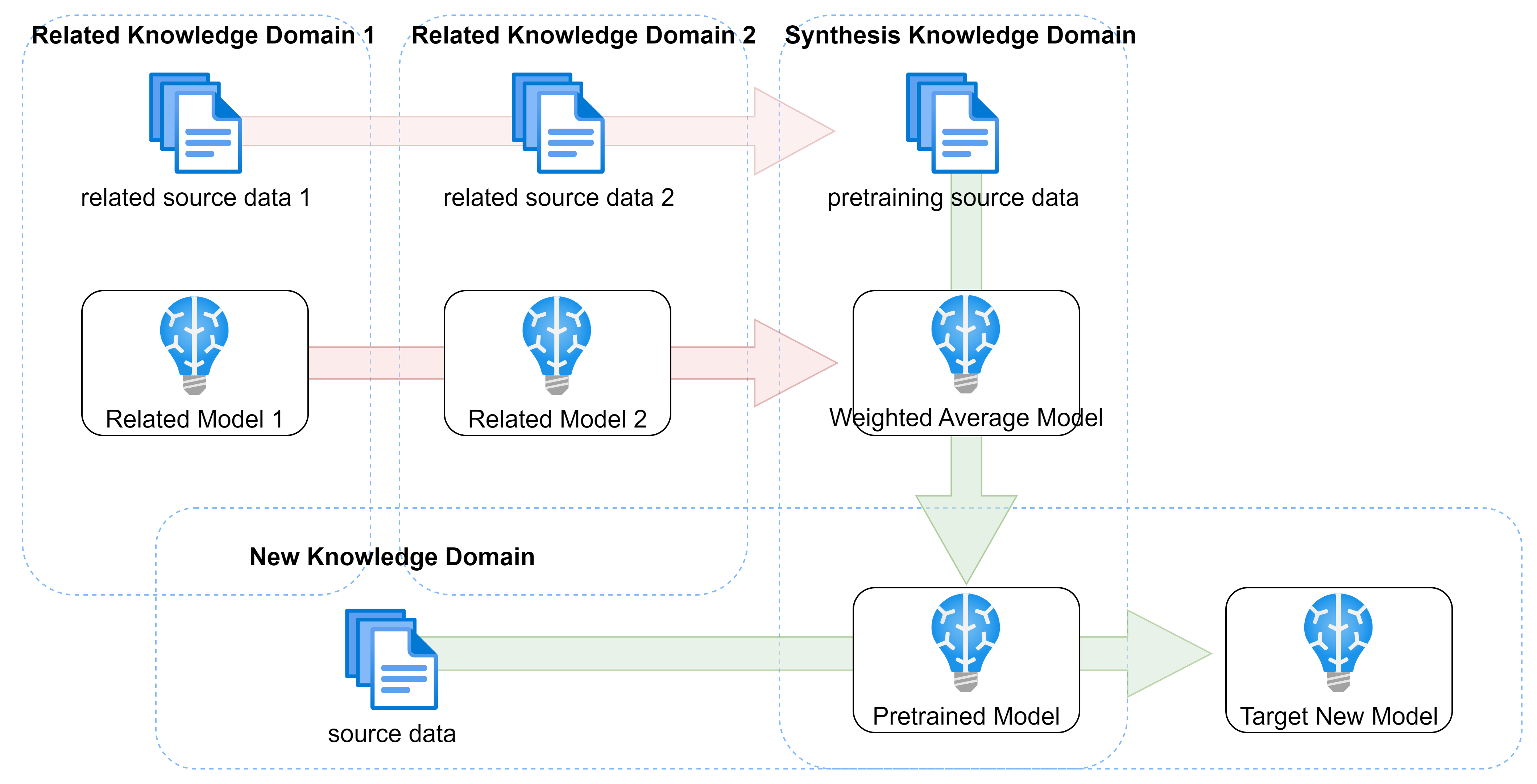}}
    \caption{Expand New Model}
    \label{fig:newmodel}
\end{figure}
When the neural network agent receives \(\textbf{F}_{\mathbf{d}^\prime}\) from the compiler and a corresponding model \(M_{\mathbf{d}^\prime}\) is not found, it indicates that a new knowledge domain has emerged. This could be due to the introduction of new files and classes, or new methods of combining multi-domain knowledge to solve problems. In such cases, the neural network agent will attempt to create a new model \(M_{\mathbf{d}^\prime}\) to learn this new knowledge, following the procedure shown in Figure~\ref{fig:newmodel}, depending on user-defined configurations in the settings file.

\subsubsection{Preliminary Model Structures}
Users can define their own strategies for choosing the structure of the neural network model through the settings file. By default, the architecture employed is a bi-directional LSTM (Long Short-Term Memory) network \cite{schuster1997bidirectional_rnn, sutskever2014sequence_lstm}.

\subsubsection{Parameter Initialization}

To initialize parameters, a set of existing knowledge domains is identified. This set, denoted as \(\mathcal{D}^{\prime}_{exist}\), comprises pre-created neural models that align with the preliminary model structures:
\begin{equation}
\mathcal{D}^{\prime}_{exist} = \{\mathbf{d}_{e,1}^{\prime}, \mathbf{d}_{e,2}^{\prime}, \ldots, \mathbf{d}_{e,n}^{\prime}\}.
\end{equation}

 The elements of this set should collectively cover the new domain's scope as comprehensively as possible and keep this set as small as possible:
 \begin{align}
 \text{Priority 1:} \quad & \min  \lvert \mathbf{d}^{\prime} \oplus (\bigcup_i \mathbf{d}_{e,i}^{\prime} \mid \mathbf{d}_{e,i}^{\prime} \in \mathcal{D}^{\prime}_{exist} ) \rvert \\
 \text{Priority 2:} \quad & \min \lvert \mathcal{D}^{\prime}_{exist} \rvert 
 \end{align}
 where \(\oplus\) denotes the symmetric difference operator.

  If \(\mathcal{D}^{\prime}_{exist}\) is empty, the Xavier method \cite{glorot2010understanding_Xavier} will be used for weight initialization. Otherwise, a shared weights strategy is applied for parameter initialization.
 
 Then, the set \(\mathcal{D}^{\prime}_{exist}\) is sorted from high to low by the Jaccard similarity coefficient \(cj\) of its element and the target knowledge domain \(\mathbf{d}^{\prime}\): 
 \begin{equation}cj = \frac{\lvert \mathbf{d}_e^{\prime} \cap \mathbf{d}^{\prime} \rvert}{\lvert \mathbf{d}_e^{\prime} \cup \mathbf{d}^{\prime} \rvert}\end{equation}
 The initial parameters \(\theta_{init}\) for \(M_{\mathbf{d}^\prime}\) are denoted as:
\begin{equation}
 \theta_{init} = \frac{\Sigma_i{cj_{\mathbf{d}^\prime_{e,i}}*\theta_{\mathbf{d}^\prime_{e,i}}}}{\Sigma_i cj_{\mathbf{d}^\prime_{e,i}}} \mid (i :\cos \langle \theta_{\mathbf{d}^\prime_{e,i}},\theta_{\mathbf{d}^\prime_{e,1}} \rangle \geq \zeta)
\end{equation}
 where \(\zeta \in [-1,1]\) serves as the cosine similarity threshold to avoid parameters that differ too much from \(\theta_{\mathbf{d}^\prime_{e,1}}\) are included in the weighted average process.

 \subsubsection{Pretraining}
After parameter initialization, the model \(M_{\mathbf{d}^{\prime}}\) undergoes a pretraining phase. During this phase, a pretraining dataset \(\mathcal{G}_{\mathbf{d}^\prime}^{pretrain}\), which is constructed by sampling from the feature sets \(\mathcal{F}_{\mathbf{d}^\prime_e}\) where \(\mathbf{d}^\prime_e \subseteq \mathbf{d}^\prime\), is utilized. This pretraining allows the network to learn the features of problem-solving strategies inherent to the contributing knowledge domains, thus mitigating the impact of potential training data scarcity. Consequently, the pretraining results in \(M_{\mathbf{d}^{\prime}}\) becoming an amalgamation of the models \(\{M_{ \mathbf{d}^{\prime}_e}\}\). However, the problems \(M_{\mathbf{d}^{\prime}}\) is now prepared to tackle may not align with those of the target knowledge domain \(\mathbf{d}^\prime\).

\subsubsection{Transfer Learning}
For \(M_{\mathbf{d}^{\prime}}\) to effectively address the grounding problems specific to the knowledge domain \(\mathbf{d}^\prime\), it undergoes further training, testing, and updating with the dataset \(\textbf{F}_{\mathbf{d}^\prime}\), following the procedures outlined in Section \ref{sec:train_test_update}.

\subsubsection{Pretraining}
After parameter initialization, a pre-training dataset $\mathcal{G}_{\mathbf{d}^\prime}^{pretrain}$ sampling from all $\mathcal{F}_{\mathbf{d}^\prime_e \mid \mathbf{d}^\prime_e \subseteq \mathbf{d}^\prime}  $ is used to enable the network learn the features of problem-solving strategies of the knowledge domains deriving it, which weakens the effect of lack of training data. After this phase, $M_{\mathbf{d}^{\prime}}$ is essentially a synthesis of $\{M_{ \mathbf{d}^{\prime}_e}\}$, therefore the grounding problems it can tackle are not perfectly consistent with the target knowledge domain $\mathbf{d}^\prime$.

\subsubsection{Transfer Learning}
To solve grounding problems of knowledge domain $\mathbf{d}^\prime$, the $M_{\mathbf{d}^{\prime}}$ is then performed to further training, testing and updating with \(\textbf{F}_{\mathbf{d}^\prime}\) just as section \ref{sec:train_test_update}.

\subsection{Policy-Making Process }
\begin{figure}[!t]
    \centering
    \makebox[\textwidth][c]{\includegraphics[width=1\linewidth]{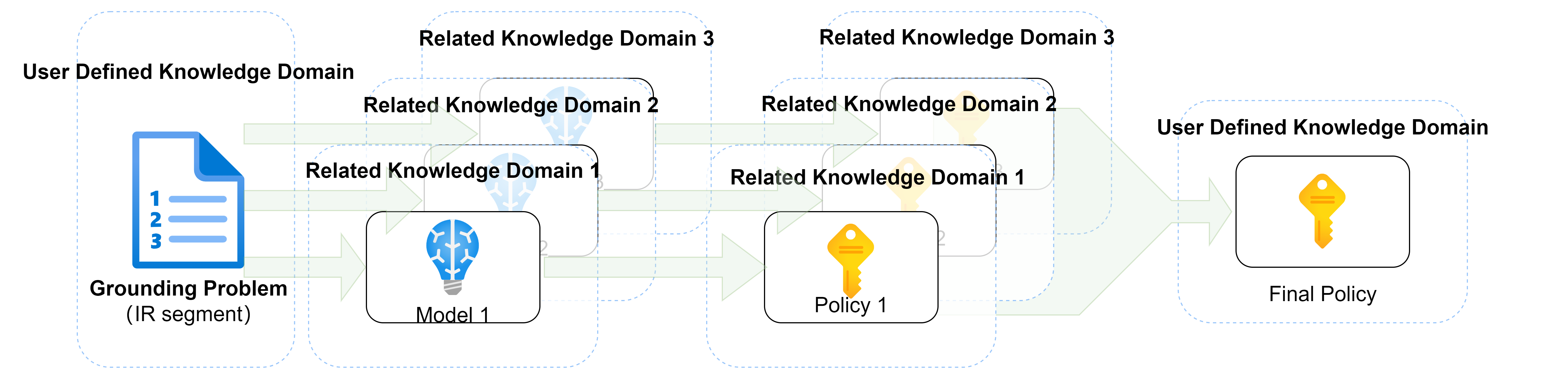}}
    \caption{Policy-Making Process}
    \label{fig:prediction}
\end{figure}
The neural network agent predicts by providing policy array $\Pi$, in domain confidence $ci$, and accuracy $ca$ to the compiler. As shown in figure \ref{fig:prediction}, the policy made by the neural network agent is the result of collaboration from models of different knowledge domains. The critical phase unfolds as follows:

\subsubsection{Selecting Collaborating Models}
When the neural network receives the grounding state $s$ and knowledge domain $\textbf{d}$ as defined by domain specification prompts (DSPs), it identifies a subset of knowledge domains with pre-established neural models, denoted as $\mathcal{D}^{\prime}{co} = {\mathbf{d}{co,0}^{\prime}, \mathbf{d}{co,1}^{\prime}, \dots, \mathbf{d}{co,n}^{\prime}}$, for the policy-making process. The union of these models should cover the full scope of $d$ effectively, ensuring that each model is a subset of $d$ and that the collective set of models is kept as small as possible:

 \begin{align}
 \text{Priority 1:} \quad & \min  \mid \mathbf{d} \setminus (\bigcup_i \mathbf{d}_{co,i}^{\prime} \mid \mathbf{d}_{co,i}^{\prime} \in \mathcal{D}^{\prime}_{co} ) \mid \\
 \text{Priority 2:} \quad & \max  \frac{\mid\{ \mathbf{d}_{co,i} \mid \mathbf{d}_{co,i}^{\prime} \in \mathcal{D}^{\prime}_{co} \text{ and } \mathbf{d}_{co,i}^{\prime} \subseteq \mathbf{d}\} \mid}{ \mid \mathcal{D}^{\prime}_{co}\mid }    \\ 
 \text{Priority 3:} \quad & \min \mid \mathcal{D}^{\prime}_{co} \mid 
 \end{align}\nonumber

  If $\mathbf{d}^{\prime}_{co}$ is empty, the neural network agent will directly return an empty policy array with both $ac$ and $ci$ set to zero. Otherwise, $\mathcal{M}_{\mathbf{d}_{co}^{\prime}}$ , which contains models corresponding to the elements of $\mathbf{D}^{\prime}_{co}$, will be loaded for prediction.

\subsubsection{Predicting and Eliminating the Outliers}

 In this phase, each element in $\mathcal{M}_{\mathbf{d}_{co}^{\prime}}$ makes 
 prediction and both result and related coefficients are stored in \begin{equation}
 \mathbf{\Pi} = \{(\mathbf{d}^{\prime}_{co,i},\Pi_i,A_{\Pi,i}, InDom_i, A_{InDom,i})\mid i \in \mathbb{N}, i < \mid \mathcal{M}_{\mathbf{d}_{co}^{\prime}} \mid \}
 \end{equation}
 Then eliminate the outliers by following the steps:
 
1  Sort $\mathbf{\Pi}$ based on  $InDom * A_{InDom}$ from high to low (in descending order), conserve the top $\eta \in (0,1]$ fraction elements. This step is used to eliminate the result generated by models not suitable to make policy for such kind of grounding state indicated by $InDom$.

2 Calculate weighted average policy array: 
\begin{equation}    
\overline{\Pi} =  \frac{\displaystyle\sum_{i=1}^{\mid \mathbf{\Pi} \mid} (A_{\Pi,i}* InDom_i* A_{InDom,i})*\Pi_i}{\displaystyle\sum_{i=1}^{\mid \mathbf{\Pi}\mid} (A_{\Pi,i}* InDom_i* A_{InDom,i})}
\end{equation}

3 then, for each element in $\mathbf{\Pi}$, if $\mathbf{d}_{co,i}^{\prime} \not\subseteq \mathbf{d}$, calculate the Symmetric Kullback-Leibler divergence between $\Pi_i $ and $\overline{\Pi}$: 
\begin{equation}
    \text{Symmetric KL}(\Pi_i, \overline{\Pi}) = \frac{1}{2} \left( \operatorname{D_{KL}}(\Pi_i \| \overline{\Pi}) + \operatorname{D_{KL}}( \overline{\Pi}\|\Pi_i ) \right)
\end{equation}
where $\operatorname{D_{KL}(p,q)}$ means the Kullback-Leibler divergence \cite{hershey2007approximating_dkl} between distributions $p$ and $q$, and eliminate the elements satisfying $\text{Symmetric KL}(\Pi_i, \overline{\Pi}) > SKL_{max}$, where $SKL_{max}$ is the threshold of the tolerant Symmetric Kullback-Leibler divergence. Those elements beyond this constraint are highly likely to misguide the compiler to apply or bind an uninvokable function belonging to knowledge domain $\mathbf{d}^{\prime}_{co,i} \setminus \mathbf{d} $.
Finally, denote the ultimate $\mathbf{\Pi}$ with no outliers as $\mathbf{\Pi}^*$

\subsubsection{Predicting and Eliminating the Outliers}

\hspace{5mm}In this phase, each element within $\mathcal{M}_{\mathbf{d}_{co}^{\prime}}$ makes prediction. The predicted policy arrays and related coefficients are stored in
\begin{equation}
\mathbf{\Pi} = \{(\mathbf{d}^{\prime}_{co,i}, \Pi_i, A_{\Pi,i}, InDom_i, A_{InDom,i}) \mid i \in \mathbb{N}, i < |\mathcal{M}_{\mathbf{d}_{co}^{\prime}}| \}
\end{equation}
Subsequently, outliers are eliminated by following these steps:

\begin{enumerate}
\item Sort $\mathbf{\Pi}$ based on the product $InDom_i* A_{InDom,i}$ in descending order and retain the top $\eta \in (0,1]$ fraction of elements. This step filters out results from models that are not adequately suited for making policy for state $s$ outside their knowledge domain, as indicated by $InDom_i$.

\item Compute the weighted average policy array
\begin{equation}
\overline{\Pi} = \frac{\sum_{i=1}^{|\mathbf{\Pi}|} (A_{\Pi,i} * InDom_i * A_{InDom,i}) * \Pi_i}{\sum_{i=1}^{|\mathbf{\Pi}|} (A_{\Pi,i} * InDom_i * A_{InDom,i})}
\end{equation}

\item For each element in $\mathbf{\Pi}$, if $\mathbf{d}_{co,i}^{\prime} \not\subseteq \mathbf{d}$, calculate the Symmetric Kullback-Leibler divergence between $\Pi_i$ and $\overline{\Pi}$:
\begin{equation}
\text{Symmetric KL}(\Pi_i, \overline{\Pi}) = \frac{1}{2} \left( \operatorname{D_{KL}}(\Pi_i \| \overline{\Pi}) + \operatorname{D_{KL}}(\overline{\Pi} \| \Pi_i) \right)
\end{equation}
where $\operatorname{D_{KL}(p \| q)}$ denotes the Kullback-Leibler divergence between distributions $p$ and $q$\cite{hershey2007approximating_dkl}. Eliminate elements where $\text{Symmetric KL}(\Pi_i, \overline{\Pi}) > SKL_{max}$, with $SKL_{max}$ representing the maximum tolerable Symmetric Kullback-Leibler divergence. Elements exceeding this threshold are likely to misdirect the compiler into applying or binding an inapplicable function belonging to the knowledge domain $\mathbf{d}^{\prime}_{co,i} \setminus \mathbf{d}$.
\end{enumerate}
Denote the ultimate $\mathbf{\Pi}$ with no outliers as $\mathbf{\Pi}^*$.

\subsubsection{Synthesizing the Policy}
\hspace{5mm}
Once outliers have been removed, synthesize the final reply policy and coefficients based on $\mathbf{\Pi}^*$:
\begin{equation}
\begin{cases}    
\begin{aligned}
    \Pi^* &=  \frac{\displaystyle\sum_{i=1}^{\mid \mathbf{\Pi}^* \mid} (A_{\Pi,i}* InDom_i* A_{InDom,i})*\Pi_i}{\displaystyle\sum_{i=1}^{\mid \mathbf{\Pi}^*\mid} (A_{\Pi,i}* InDom_i* A_{InDom,i})} \\
ci^* &= \frac{\displaystyle\sum_{i=1}^{\mid \mathbf{\Pi}^* \mid} (InDom_i* A_{InDom,i})}{\displaystyle\sum_{i=1}^{\mid \mathbf{\Pi}^*\mid} (A_{InDom,i})}\\
ac^* &= \frac{\displaystyle\sum_{i=1}^{\mid \mathbf{\Pi}^* \mid} (A_{\Pi,i}* InDom_i* A_{InDom,i})}{\displaystyle\sum_{i=1}^{\mid \mathbf{\Pi}^*\mid} (InDom_i* A_{InDom,i})}\\
\end{aligned}
\end{cases} 
\end{equation}
where $(\mathbf{d}^{\prime}_{co,i},\Pi_i,A_{\Pi,i}, InDom_i, A_{InDom,i}) \in \mathbf{\Pi}^*$.

Send the $(\Pi^*,ac^*,ci^*)$ to the compiler to guide the grounding process together with process control prompts (PCPs).

\section{Advanced Research Directions}
\hspace{5mm}
In the era of rapid development of artificial intelligence, the COOL and relative concepts proposed in this article may have outstanding prospects in the following aspects:
\subsection{Infrastructure for Edge Computing and Federated Learning}
The COOL, and the U2N mechanism, are intrinsically suited for edge computing and federated learning\cite{abreha2022federated_fl_ec}. As edge devices proliferate, localized data processing and decision-making become vital. By augmenting COOL's network support for edge environments, it can serve as a platform that facilitates seamless user interaction with edge computing and federated learning ecosystems. This integration is promising as it addresses pivotal concerns such as data privacy and the hinder of model training and iterative updates. Moreover, it enhance the inter-device collaboration, thereby enhancing the overall intelligence of distributed networks.

\subsection{Mitigating the Dependency on Large Language Models' Scale through Neural-Symbolic Integration}
\begin{figure}[!h]
    \centering
    \includegraphics[width=0.7\linewidth]{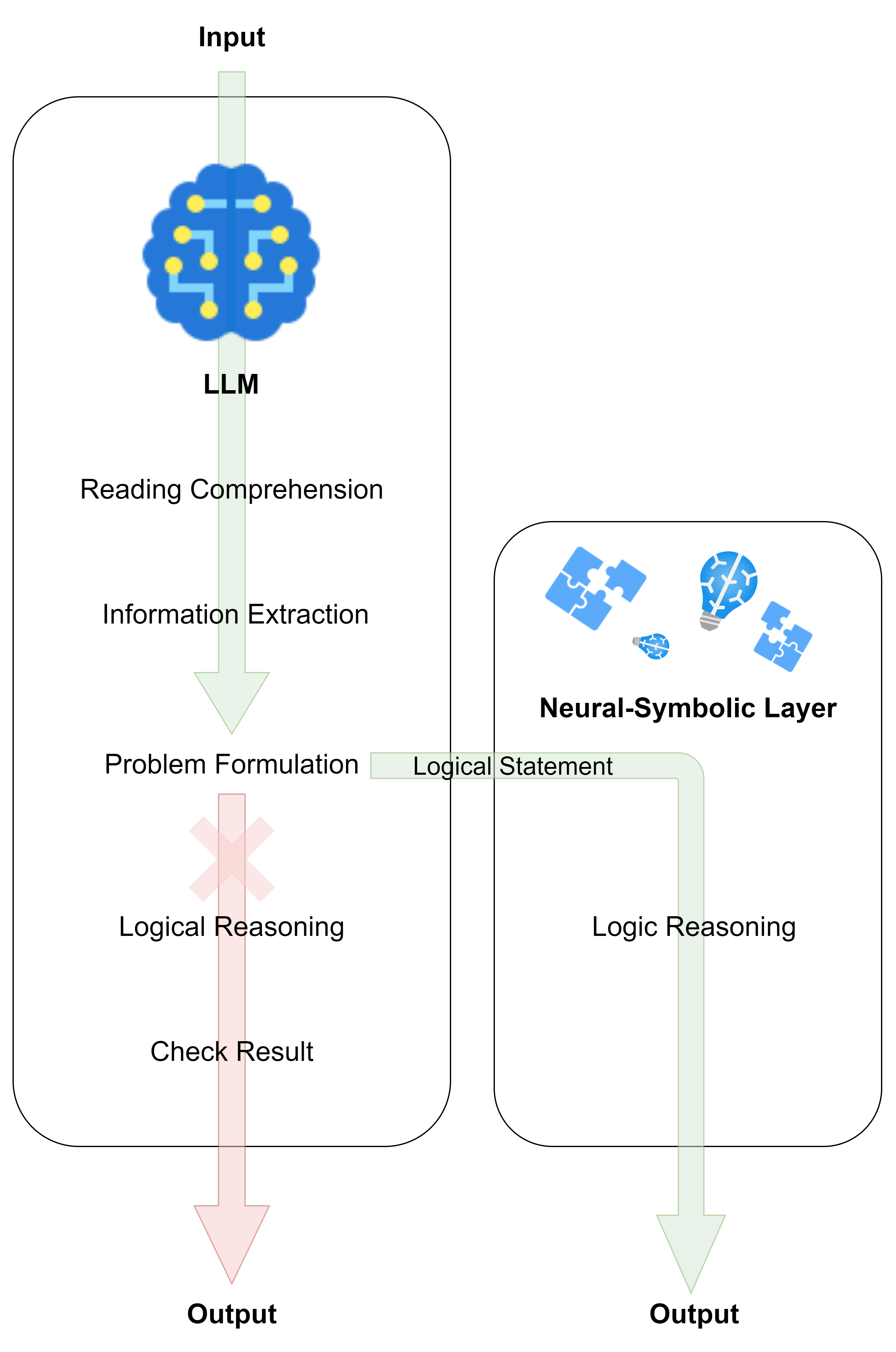}
    \caption{LLM with a Neural-Symbolic Layer}
    \label{fig:nsl-llm}
\end{figure}
Built upon the Transformer architecture \cite{vaswani2017attention_transformer}, Large Language Models (LLMs) have shown remarkable performance in natural language processing \cite{min2023recent_natural_languageLLM}  and complex reasoning \cite{huang2022towards_LLM_inReasoning}. Despite their impressive capabilities, these models exhibit a concerning trend: their reasoning capacity is heavily reliant on the scale of the model, this relationship is characterized by rapidly diminishing returns. Consequently, each incremental improvement in reasoning ability results in considerably higher resource consumption compared to previous advancements. This pattern of growth is unsustainable, presenting both a developmental bottleneck and an environmental challenge for the ongoing deployment of LLMs.

As depicted in the Figure \ref{fig:nsl-llm}, a novel approach to this challenge involves the incorporation of neural-logic layers into the LLM framework to replace the logical reasoning process. In this configuration, the LLM specializes in tasks such as reading comprehension, information extraction, and problem formulation. Subsequently, the reasoning process is handled by the neural-symbolic layers, which are adept at performing logical reasoning tasks and have strong scalability. This layered structure promises to alleviate the constraints imposed by the physical scale of LLM networks, offering a pathway to enhance inferential capabilities without the associated resource intensity of traditional scaling methods.



\section{ Key Implementation Details}
\subsection{Hierarchical Address}
\hspace{5mm}
In the IR segment,  a hierarchical structure is adopted to denote the addresses in the IR. \begin{equation}
A = \{a_1, a_2, \ldots, a_n\}, \text{ where } a_i \in \mathcal{N} \text{ and } 1 \leq i \leq n
\end{equation}
This allows any number of addresses between the two addresses, ensuring that when the IR is partially replaced in the grounding process, the newly inserted IR segment always has enough addresses while not modifying other addresses.

\subsection{Model Pool}
\hspace{5mm}
 For efficient storage and reuse of neural network models, a model pool with a Least Recently Used (LRU) strategy \cite{lee2001lrfu_lru} is employed. As the model pool reaches its capacity, this strategy helps to identify and discard the least accessed models, thus prioritizing the retention of the most relevant and frequently used models. Moreover, a protection mechanism is in place for models that have been recently added to the pool, preventing their premature elimination and affording them a grace period to demonstrate their utility and relevance. These provisions are designed to optimize the use of computational resources.

\section{Design of Experiment}
\subsection{Basic Experiments}
\subsubsection{Effect of Process Control Prompt}
\begin{itemize}
    \item \textbf{Purpose:} To evaluate the impact of Process Control Prompts (PCPs) on the grounding process, and to observe how PCPs influence the quality of the generated Intermediate Representation (IR) as reflected by the execution speed of the resulting compiled program.
    \item \textbf{Control Variables:} Grounding algorithm (BDDB), neural network agent disabled.
    \item \textbf{Independent Variables:} PCPs enabled, PCPs disabled (with a setting that causes BDDB to operate like Breadth-First Search (BFS) ), PCPs disabled (with a setting that causes BDDB to operate like Depth-First Search (DFS) )
    \item \textbf{Dependent Variables:} Success rate of compilation ($p_{suc}$), the average number of grounding states generated, average time taken for grounding and average execution time of the compiled program.

    \item  \textbf{Materials:} COOL source code for solving math problems from \href{https://math-drills.com}{https://math-drills.com} and similar sites. 
    \item  \textbf{Procedure:} Compile the COOL source code with different configurations and execute the resulting program.
\end{itemize}

\subsubsection{Effect of BDDB algorithm}
\begin{itemize}
    \item \textbf{Purpose:} To evaluate the effect of BDDB algorithm on the grounding process and enhancing the quality of the generated IR.
    \item \textbf{Independent Variables:} BDDB algorithm (with $\gamma = 0$ and $\lambda = 0$, which simplifies the approach to an approximation of greedy-policy Markov Decision Processes (MDPs) with the capability of backtracking), BDDB algorithm (with $\gamma = 0$ and a fine-tuned $\lambda$)
    \item \textbf{Dependent Variables:} Success rate of compilation ($p_{suc}$), the average number of grounding states generated, average time taken for grounding and average execution time of the compiled program.

    \item  \textbf{Materials:} COOL source code for solving math problems. 
    \item  \textbf{Procedure:} Compile the COOL source code with different configurations and execute the resulting program.
\end{itemize}

\subsubsection{Effect of U2N on Single Knowledge Domain Problem Solving}
\begin{itemize}
    \item \textbf{Purpose:} To assess the U2N algorithm's impact on the grounding process and the quality of IR for single-domain problem-solving.
    \item \textbf{Control Variables:} Neural network agent (prediction enabled and initially without prior training), BDDB (with $\gamma = 0$ and a fine-tuned $\lambda$).
    \item \textbf{Independent Variables:} The quantity of different COOL source codes of compiled with the compiler with neural network agent learning mode enabled.
    \item \textbf{Dependent Variables:} The accuracy of the neural network's predictions, the success rate of compilation (\(p_{suc}\)), the average number of grounding states generated, and the average grounding and average execution times post-compilation.

    \item  \textbf{Materials:} COOL source code samples that address math problems within a specific knowledge domain.
    \item  \textbf{Procedure:} Compile the COOL source code with the neural network agent's learning mode enabled and subsequently, for evaluation purposes, with the learning mode disabled.
    \item  \textbf{Analysis:} Draw graphs of the dependent variables as functions of the independent variable to evaluate the learning ability of the neural network agent.
    and draw graphs of (the percentage of compilation completed successfully, average number of grounding states created, and average grounding time as functions of the accuracy of neural network models, and average program execution time after compilation) to evaluate the effect of neural network agents taking a bigger and bigger proportion in policy making.
    Generate graphical representations of the dependent variables in relation to the independent variable to analyze the learning efficacy of the neural network agent. Also, graph the relationship between neural network model's accuracy (policy-making influence) in terms of successful compilation rates, grounding states, and execution timings.
\end{itemize}

\subsubsection{Effect of U2N on Multi Knowledge Domain Problem Solving}
\begin{itemize}
    \item \textbf{Purpose:} To evaluate the capability of the U2N mechanism to handle multiple knowledge domains during the grounding process and to assess the quality of the generated IR.
    \item \textbf{Control Variables:} BDDB algorithm (with $\gamma = 0$ and a fine-tuned $\lambda$).
    \item \textbf{Independent Variables:} Neural network agent (Predicting enabled, learning mode disabled, well trained for handling problems of knowledge domain $\mathbf{d}_1^{\prime}$ and $\mathbf{d}_2^{\prime}$), Neural network agent disabled, different knowledge domains in the COOL source code being compiled.
    
    \item \textbf{Dependent Variables:} The accuracy of neural network predictions, the success rate of compilation, the average number of grounding states generated, and the average grounding time.

    \item  \textbf{Materials:} COOL source code representing mathematical problems from the combined knowledge domains \(\mathbf{d}_1\) and \(\mathbf{d}_2\), where \(\mathbf{d}_1\) includes domains \(\mathbf{d}_1^{\prime}\) and \(\mathbf{d}_2^{\prime}\), and \(\mathbf{d}_2\) includes \(\mathbf{d}_1^{\prime}\), \(\mathbf{d}_2^{\prime}\), and \(\mathbf{d}_3^{\prime}\).
    
    \item  \textbf{Procedure:} Sequentially compile the COOL source code with the neural network agent both enabled and disabled.

\end{itemize}
\subsection{Advanced Experiments}
\subsubsection{Effect of combination of LLM with neural-symbolic layer}
\begin{itemize}
    \item \textbf{Background and Purpose:} As the increase in model size for large language models (LLMs) such as GPT series\cite{radford2018improving_GPT1} leads to diminishing returns, this experiment seeks to determine whether combining a smaller LLM with a neural-symbolic layer can match or surpass the logical reasoning abilities of a larger LLM alone, providing an alternative approach for enhancing AI reasoning.
    \item \textbf{Independent Variables:} Comparison of GPT-2\cite{radford2019language_gpt2} and GPT-3\cite{brown2020language_GPT3}, each augmented with a neural-symbolic layer, against a baseline version of a larger GPT model (e.g., GPT-4)
    \item \textbf{Dependent Variables:} The reasoning accuracy, efficiency, and complexity of problems that can be solved by each framework.
    \item  \textbf{Materials:} Math word problems varying in complexity.
    \item  \textbf{Procedure:} Conduct problem-solving sessions using each framework to solve the dataset of math word problems.
    
\end{itemize}

\subsubsection{Comparison of reasoning ability of different LLM-based frameworks}
\begin{itemize}
    \item \textbf{Background:} Research like VisProg \cite{gupta2023visual_prog} and Program-of-Thoughts\cite{chen2022program_CoP} has shown improvements in reasoning ability when LLMs are used to generate Python code. However, these benefits often rely on the LLM's ability to formulate a correct Chain-of-Thought \cite{wei2022chain_thought_chain_CoT} (CoT). Complex problems that exceed an LLM's reasoning capacity may be unsolvable within this paradigm. Such limitations may be overcome by integrating a neural-symbolic layer, which can take over the complex logical reasoning process after the problem formulation stage, as illustrated in Figure \ref{fig:nsl-llm}.
    \item \textbf{Purpose:} To evaluate how adding a neural-symbolic layer to an LLM affects its reasoning capabilities, especially for complex problems where chain-of-thought approaches may fail.
    \item \textbf{Control Variables:} Consistent GPT version across all frameworks.
    \item \textbf{Independent Variables:} Three frameworks are compared: GPT augmented with chain-of-thought prompting (CoT), GPT enhanced with chain-of-prompts (CoP), and GPT integrated with a neural-symbolic layer. 
    \item \textbf{Dependent Variables:} The accuracy and efficiency of problem-solving.
    \item  \textbf{Materials:} Math word problems varying in complexity.
    \item  \textbf{Procedure:} Carry out problem-solving tasks using each framework to solve the dataset of math word problems.
    
\end{itemize}

\subsection{Long-Term Experiment}
\subsubsection{Application of COOL}
\begin{itemize}
    \item \textbf{Purpose:} To evaluate the COOL programming language's readiness for production use by assessing its features, performance, and user satisfaction over a series of tests.
    \item \textbf{Independent Variables:} Language features, algorithm optimization, execution environment, documentation.
    \item \textbf{Dependent Variables:}  Application areas, performance, user feedback.
\end{itemize}


\bibliography{main}

\section{*About the Author}
\hspace{5mm}

We are exploring new research areas and rely on thorough experiments to support our work. We've started detailed experiments to test and improve our proposed ideas.

These experiments need significant computing power and expertise. We see the value in our work and know collaboration can enhance its impact.

I'm looking for a Ph.D. program that aligns with my research interests.

Meanwhile, we welcome the community's input, advice, and partnership offers to advance our field together.

\end{document}